\documentclass[11pt]{article}

\PassOptionsToPackage{table}{xcolor}

\usepackage[final]{acl}

\usepackage{times}
\usepackage{latexsym}

\usepackage[T1]{fontenc}

\usepackage[utf8]{inputenc}

\usepackage{microtype}

\usepackage{inconsolata}

\usepackage{graphicx}

\usepackage{graphicx}
\usepackage{booktabs}
\usepackage{bm}
\usepackage{multirow}
\usepackage{ifthen}
\usepackage{dsfont}
\usepackage{pifont}
\usepackage{nicefrac}
\usepackage{bigstrut}
\usepackage{tabularx}
\usepackage{amsmath}
\usepackage{amssymb}
\usepackage{algorithm}
\usepackage{algpseudocode}

\newcommand{\method}[1]{CFD}

\newcommand{\cmark}{\ding{51}}%
\newcommand{\xmark}{\ding{55}}%

\def\onedot{.}
\def\eg{\emph{e.g}\onedot}

 \def\vs{\emph{vs}\onedot}

\usepackage{cleveref}

\title{Caption-once, Frames-on-Demand: Visual-Need Routing for Budget-Aware Agentic Long Video Understanding}

\author{
  \textbf{Weitong Cai}$^{1}$,
  \textbf{Hang Zhang}$^{2}$\footnotemark[1],
  \textbf{Yukai Huang}$^{3}$,
  \textbf{Yiqiao Xie}$^{4}$, \\
  \textbf{Shan Gao}$^{5}$,
  \textbf{Jiankang Deng}$^{4}$,
  \textbf{Songcen Xu}$^{5}$,
  \textbf{Jifei Song}$^{5}$,
  \textbf{Zhensong Zhang}$^{5}$\thanks{Corresponding authors.}\\
  $^{1}$Queen Mary University of London, $^{2}$Independent Researcher, \\
  $^{3}$Durham University, 
  $^{4}$Imperial College London, $^{5}$Huawei \\
  {\tt\small weitong.cai@qmul.ac.uk, miruku.hzhang@gmail.com, zhangzhensong@huawei.com}
}

\begin{document}
\maketitle

\begin{abstract}
Long-video understanding on edge devices must reason over hours of content
under tight compute and bandwidth budgets.
Subsampling visual tokens loses temporal structure, while text-only video
memories lose fine-grained visual attributes.
We observe a visual-textual duality: language memories carry long-range
temporal structure better than dense frames, while pixels remain decisive
for attribute-level perception.
Building on this insight, we propose \textbf{Caption-once, Frames-on-Demand
(\method{abbr})}, a budget-aware edge-cloud agentic framework.
The edge runs a single offline captioning pass that builds a dual-track
narrative index, an event-level story skeleton plus a clip-level
micro-log, cached and reused across queries without re-captioning.
At query time, a cloud-side MLLM reasons over the index in a story-first
loop centered on a \textbf{lightweight Visual-Need Router}: a per-query
gating module that triggers bounded keyframe retrieval only for
perceptual questions (appearance, on-screen text, attribute disambiguation)
and keeps temporal-structural questions in language space.
The router turns visual access into a first-class, query-conditioned
cost, capping per-query frame consumption regardless of video length.
Experiments on long-video benchmarks demonstrate strong
accuracy-efficiency trade-offs while substantially reducing online visual
processing.
\end{abstract}

\section{Introduction}
\label{sec:intro}

\begin{figure*}[t]
\centering
\includegraphics[width=0.9\linewidth]{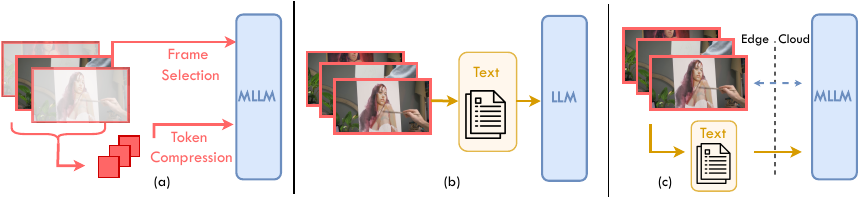}
\caption{
\textbf{Long video understanding.}
\textbf{(a)}~Visual compression methods (frame selection or token
compression) reduce visual tokens fed to an MLLM but risk losing
coverage on long videos.
\textbf{(b)}~Textual translation methods convert video into language
memories for LLM reasoning, improving scalability but losing
fine-grained visual evidence.
\textbf{(c)}~Our approach separates offline narrative indexing on the
edge from on-demand visual verification on the cloud, combining textual
temporal coverage with targeted pixel-level evidence only when needed.
}
\vspace{-3mm}
\label{fig:teaser} 
\end{figure*}

Long-form video understanding underlies a growing class of multimodal
applications, from interactive video assistants and on-device personal
memory to wearable visual aids, each constrained by tight inference budgets
imposed by either edge hardware or latency-sensitive cloud
serving~\cite{paruchuri2025egotrigger,singh2023edge}.
Under such constraints, long video QA becomes a joint modeling and
resource-allocation problem: the system must preserve query-relevant memory
over the entire video horizon, yet missing even a small but decisive event
or attribute can directly cause incorrect
answers~\cite{mangalam2023egoschema}.

Existing approaches can be viewed along two intertwined design axes:
\textbf{(i) Visual compression} methods (\cref{fig:teaser}(a)) 
reduce cost by sampling fewer frames or compressing visual tokens, but may 
sacrifice coverage or fidelity on long 
videos~\cite{qian2024streaming,jiang2025storm}.
\textbf{(ii) Textual translation} methods (\cref{fig:teaser}(b)) 
convert videos into captions, transcripts, or language memories and perform 
retrieval and reasoning in text space, or iteratively revisit video 
segments to produce updated textual memories across multiple 
rounds~\cite{fan2024videoagent,zhi2025videoagent2,zuo2025videolucy, cai2025mllm}.
These strategies improve scalability, but text-centric representations can 
lose fine-grained visual 
evidence~\cite{wang2023lifelongmemory,kahatapitiya2025language}, while 
iterative revisiting often introduces query-time cost that grows with 
interaction rounds and is difficult to bound under strict edge 
budgets~\cite{fan2024videoagent,zhi2025videoagent2}.

In preliminary studies, we observe a consistent duality.
For image-level referencing (Table~\ref{tab:pre_CapQA}), caption-only reasoning falls short of reasoning from original pixels, since fine-grained properties such as identity, color, and attribute binding are easily lost under aggressive linguistic compression~\cite{yang2025captionqa}.
In contrast, for long-horizon temporal reasoning (Table~\ref{tab:pre_infini}), language memories can sometimes be more effective than dense visual inputs for chronological understanding and event-level relations, because they provide a compact, discrete record of what happened and in what order~\cite{wang2023lifelongmemory,kahatapitiya2025language}.
This suggests that text is not merely a lossy substitute for video: it can be a stronger carrier for temporal structure, while sparse raw frames remain essential for verifiable visual evidence.
The intuition aligns with human memory: we rarely retain every pixel-level detail, but we can retain and query a concise storyline of an experience.

Motivated by this observation, we propose a \textbf{Caption-once, Frames-on-Demand (\method{abbr})} paradigm (\cref{fig:teaser}(c)) for budget-aware long video understanding in an edge-cloud setting.
The key idea is to index temporal structure once in language and reuse it across queries, while retrieving raw frames only when fine-grained visual attribution is necessary.
Under this view, efficiency does not come from uniformly reducing visual computation, but from allocating visual access selectively to attribute-critical moments.
We instantiate this paradigm with \method{}, a budget-aware edge-cloud agentic pipeline for long video understanding.
On the edge, a lightweight MLLM performs \emph{single-pass, question-agnostic indexing} of the full video stream to build two complementary timestamped language memories: an \emph{event-level story skeleton} derived from shot/event segmentation~\cite{soucek2024transnet}, and a \emph{uniform clip-level micro-log} that preserves local temporal details between events.
Both memory tracks are produced once per video and cached for reuse across subsequent queries without re-captioning the video.

\begin{table*}[t]  %
\begin{minipage}[t]{0.3\linewidth}
\centering
\footnotesize
\caption{CaptionQA~\cite{yang2025captionqa}. Caption:\texttt{Qwen2.5-VL-7B}; Reasoning:\texttt{Qwen3-VL-4B}.}
\resizebox{0.8\linewidth}{!}{%
\begin{tabular}{lc}
\hline
Input & Overall (\%) \bigstrut\\
\hline
\hline
Image & 92.32 \bigstrut[t]\\
Caption & 77.17 \bigstrut[b]\\
\hline
\end{tabular}%

}
\label{tab:pre_CapQA}
\end{minipage}
\hfill
\begin{minipage}[t]{0.67\linewidth}
\footnotesize
\centering
\caption{InfiniBench~\cite{ataallah2025infinibench}. Use \texttt{Qwen3-VL-32B} for caption and reasoning. Results show captions have better performance on temporal context understanding than raw videos. }
\renewcommand{\arraystretch}{0.85}  %
\resizebox{\linewidth}{!}{%
\begin{tabular}{lcccc}
\hline
\multirow{2}[2]{*}{Pipeline} & Chronological  & Global  & Scene  & Character  \bigstrut[t]\\
      & Understanding  & Appearance & Transitions & Actions \bigstrut[b]\\
\hline
\hline
Video & 48.44 & 70.54 & 53.97 & 67.04 \bigstrut[t]\\
60s / Caption & 56.46 & 53.49 & 58.73 & 53.07 \bigstrut[b]\\
\hline
\end{tabular}%

}
\label{tab:pre_infini}
\end{minipage}
\end{table*}

At query time, a stronger cloud-side MLLM reasons over the cached narrative index in an iterative backtracking loop.
It first attempts a \emph{story-first} answer from the event memory.
If confidence is insufficient, the agent \emph{localizes} a relevant unexplored event, \emph{enriches} it with overlapping clip-level captions, and re-attempts answering.
A lightweight \emph{Visual-Need Router} then predicts whether pixel-level inspection is required: it favors frame retrieval for perceptual queries (\eg, appearance, on-screen text) and skips it for temporally grounded questions when narrative memory is sufficient.
When visual inspection is triggered, the system retrieves a bounded set of keyframes from the localized event into a fixed-capacity FIFO working memory, ensuring that per-query visual cost remains controlled regardless of video length.
The agent then performs \emph{multimodal re-reasoning} over textual memory and retrieved visual evidence in a unified prompt.
The loop repeats until confidence is reached or the query budget is exhausted.

In summary, our contributions are threefold:
\begin{itemize}
  \item \textbf{Visual-Need Router.} We introduce a per-query gating module that classifies each question as perceptual or temporal-structural and triggers raw-frame retrieval only for the former. To our knowledge, this is the first agentic video-QA design that conditions visual access on question type, converting frame retrieval from an emergent pipeline byproduct into an \emph{explicit, query-conditioned cost}.
  \item \textbf{Caption-once, Frames-on-Demand framework.} Building on prior caption-once paradigms~\cite{zhang2024simple,kahatapitiya2025language}, we couple a reusable dual-track narrative index (event skeleton plus clip-level micro-log) with the Visual-Need Router and a fixed-capacity FIFO working memory, exposing per-query visual cost as an explicit hyperparameter rather than an implicit consequence of interaction depth.
  \item \textbf{Empirical study.} On Video-MME and InfiniBench, \method{abbr} reaches accuracy competitive with prior agent-based methods while using roughly an order of magnitude fewer frames per question, supporting the design hypothesis that visual access can be allocated selectively rather than uniformly.
\end{itemize}

\noindent An extended discussion of related work is provided in Appendix~\ref{sec:related_work}.

\section{Method}
\label{sec:method}

\begin{figure*}[t]
\centering
\includegraphics[width=0.75\textwidth]{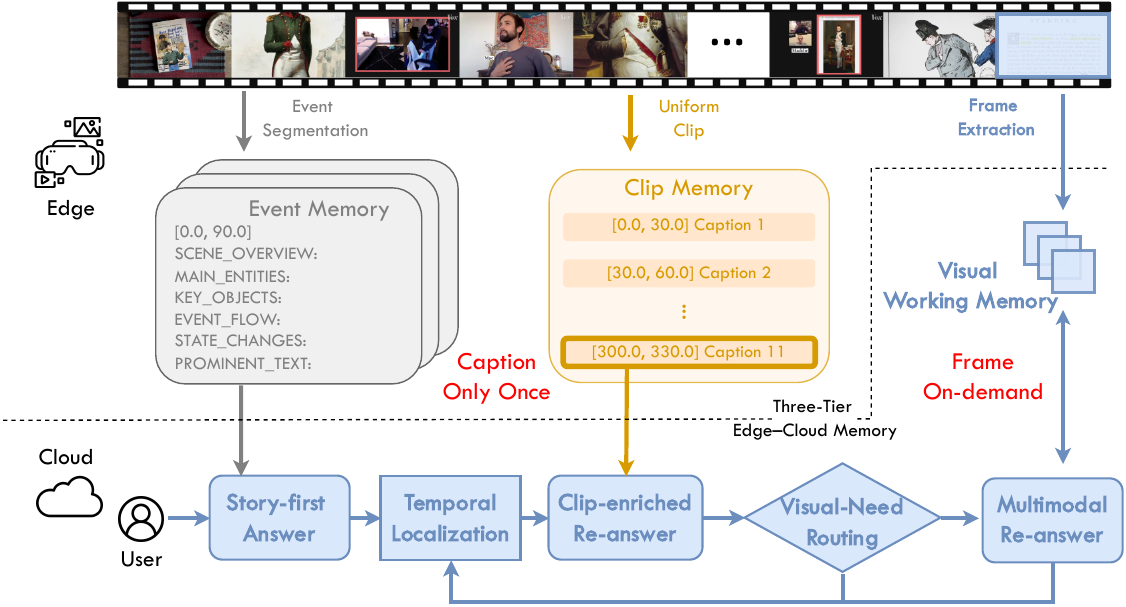}
\caption{
\textbf{Overview of \protect\method{}.}
On the edge, a lightweight MLLM performs a single offline captioning pass to build
a dual-track narrative index: an event-level story skeleton and a clip-level micro-log,
both cached and reused across queries without re-captioning.
At query time, a cloud-side MLLM reasons over the cached index through a story-first
iterative loop, enriching targeted events with clip details and retrieving raw frames
on demand only when the Visual-Need Router determines pixel-level evidence is necessary.
}
\label{fig:framework} 
\end{figure*}

We present Caption-once, Frames-on-Demand (\method{abbr}), a budget-aware edge-cloud agentic framework for long video understanding.
The framework consists of three parts: a three-tier memory architecture that separates offline narrative indexing from online visual verification (\cref{sec:memory}), four agent roles (\cref{sec:agents}), and a story-first reasoning loop that orchestrates these agents under a controlled visual budget (\cref{sec:loop}).
An overview is shown in \cref{fig:framework}.

\subsection{Three-Tier Edge-Cloud Memory Architecture}
\label{sec:memory}

A central challenge in long video understanding is that different questions 
demand different evidence.
Temporal and causal questions (\eg, ``What happened after the man left?'') 
can often be resolved from a compact narrative record, while perceptual 
questions (\eg, ``What color was the mug?'') require direct pixel-level 
verification.
Serving both needs efficiently calls for a memory that provides broad 
temporal coverage by default, fine-grained textual detail on demand, and 
targeted visual evidence only when language alone is insufficient.

\method{abbr} addresses this with a three-tier memory architecture under an edge-cloud setting.
Prior agent-based systems~\cite{zuo2025videolucy} adopt a hierarchical memory where deeper levels are generated \emph{online} by repeatedly invoking the captioning model during question answering, incurring query-time visual cost that grows with interaction depth.
In contrast, \method{abbr} constructs all textual memories \emph{offline} in a single question-agnostic indexing pass, and introduces a third, \emph{visual} tier that provides raw pixel evidence when text is insufficient.
This decoupling ensures that query-time reasoning operates primarily in language space, with visual access as a bounded, selective supplement rather than a recurring expense.
We emphasise that the framework does not presume the offline captions to be
sufficient on their own: the visual tier is designed precisely for cases
where the question-agnostic index lacks the needed detail, with the
Visual-Need Router (\cref{sec:agents}) deciding when to escalate from text
to pixels rather than committing to a possibly wrong text-only answer.

\noindent\textbf{Tier~1: Event memory (global story skeleton).}
Given a video $V$ of duration $D$ seconds, we segment it into semantically coherent events using a shot-boundary detector~\cite{soucek2024transnet}.
Adjacent segments shorter than a threshold $\tau_{\min}$ are merged into the preceding segment, yielding variable-length events $\{e_1, \ldots, e_K\}$. 
A lightweight captioning MLLM then processes each event to produce a structured narrative profile:
\begin{equation}
  m_k^{E} = \text{Captioner}(e_k,\; p_{\text{event}}),
  \label{eq:event_mem}
\end{equation}
where a prompt $p_{\text{event}}$ instructs the captioner to produce a story-skeleton 
description including scene overview, entity listing, chronological event 
flow, and retrieval anchor tags.
The full event memory 
$\mathcal{M}_E = \{m_k^{E}\}_{k=1}^{K}$ 
spans the entire video and is \emph{always presented in full} to the 
reasoning agent, serving as the global timeline and starting point for all 
question answering.

\noindent\textbf{Tier~2: Clip memory (local temporal details).}
Independent of event segmentation, we partition $V$ into fixed-length clips of duration $\tau_c$ with configurable stride, yielding a sequence of clips $\{c_1, \ldots, c_L\}$.
Each clip is captioned with a micro-action-level prompt $p_{\text{clip}}$ designed for high temporal granularity, one action per sentence:
\begin{equation}
  m_j^{C} = \text{Captioner}(c_j,\; p_{\text{clip}}).
  \label{eq:clip_mem}
\end{equation}
$\mathcal{M}_C=\{ m_j^{C}\}_{j=1}^{L}$ captures local temporal details (micro-actions, state changes, visible text) that may be too fine-grained for the event-level skeleton.

Crucially, the clip memory is \emph{not} fed to the reasoning agent in full.
Instead, it is activated selectively: when the reasoning loop localizes a specific event $e^*$, all clip captions whose time intervals overlap with $e^*$ are injected as nested sub-descriptions within $e^*$'s narrative text (\cref{sec:loop}).
This on-demand injection avoids overwhelming the LLM context with full clip-level detail for the entire video, while providing dense local information exactly where it is needed.
Moreover, these enrichments are \emph{cumulative}: once clip captions are injected for an explored event, they persist in the narrative memory for all subsequent reasoning steps within the same query.
This mechanism replaces the online re-captioning used in prior work~\cite{zuo2025videolucy}: instead of invoking the MLLM again to generate finer descriptions, \method{abbr} composes already-cached captions through pure text operations at zero additional visual cost.

\noindent\textbf{Tier~3: Visual working memory (pixel-level evidence).}
The first two tiers operate entirely in language space.
For questions that require direct visual verification, \method{abbr} maintains a Visual Working Memory $\mathcal{W}$, a FIFO queue with a deliberately small fixed capacity $N_{\text{wm}}$ that stores raw image frames extracted from the video.
When the reasoning loop determines that pixel-level inspection is necessary for an event $e^*$, keyframes are adaptively sampled from $e^*$ and pushed into $\mathcal{W}$; when $\mathcal{W}$ is full, the oldest frames are evicted.
The capacity $N_{\text{wm}}$ is set to be orders of magnitude smaller than the total frame count of the video (e.g., 16 frames for a one-hour video).
This compact design serves two purposes: it prevents excessive visual tokens from diluting the LLM's attention over the context, and it minimizes the volume of visual data transmitted from edge to cloud, keeping uplink bandwidth and energy consumption under control.
By introducing a bounded visual channel alongside the narrative, \method{abbr} can ground perceptual claims in actual pixels while keeping the visual footprint minimal.

The three tiers form a coarse-to-fine progression that mirrors the caption-once, frames-on-demand principle: Event Memory provides breadth across the full video, Clip Memory adds depth at targeted time periods, and Visual Working Memory supplies pixel-level evidence when language alone is insufficient.
All three tiers are jointly presented to the answering agent in a single multimodal prompt, with textual memories providing the reasoning context and visual evidence grounding fine-grained attributes.

\subsection{Agent Roles in the \protect\method{} Framework}
\label{sec:agents}

\method{abbr} employs four agent roles instantiated through prompt engineering, inspired by prior multi-agent designs for video understanding~\cite{zuo2025videolucy,zhi2025videoagent2,fan2024videoagent}.
A key difference from these systems is the \emph{absence of an Instruction Agent}: prior pipelines use a dedicated agent to analyze what information is missing and generate re-captioning instructions, triggering additional VLM calls at query time~\cite{zuo2025videolucy}.
\method{abbr} eliminates this role entirely by substituting it with two complementary mechanisms: the pre-computed clip memory (which already contains the finer-grained descriptions that an Instruction Agent would request) and a Visual-Need Router (which decides when to look at raw pixels instead of generating more text).

\noindent\textbf{Captioning Agent (offline only).}
An edge-side MLLM processes each event clip and uniform clip according to \cref{eq:event_mem,eq:clip_mem}.
This agent is active \emph{only during the offline indexing stage} on the edge.
Once the dual-track memory is cached, the agent is not invoked again for any subsequent query on the same video.

\noindent\textbf{Answering Agent.}
Given the current narrative memory $\mathcal{M}$ (which may include injected clip details) and the visual working memory $\mathcal{W}$, this agent evaluates whether the available evidence is sufficient to answer the question $Q$ with confidence:
\begin{equation}
  (a,\; \textit{conf},\; r) = \textsc{AnsAgent}(\mathcal{M},\; \mathcal{W},\; Q),
  \label{eq:ans_agent}
\end{equation}
where $a$ is the candidate answer, $\textit{conf} \in \{\textit{true}, \textit{false}\}$ is the confidence flag, and $r$ is a chain-of-thought reason.
The textual memory and visual evidence are assembled into a single multimodal prompt: images from $\mathcal{W}$ are interleaved with their timestamps before the textual narrative, allowing the LLM to jointly ground visual details and temporal context.
When $\textit{conf} = \textit{true}$, the answer $a$ is returned; otherwise, the system enters/continues the reasoning loop.

\noindent\textbf{Localization Agent.}
This agent identifies the single most question-relevant event from the narrative memory, excluding previously explored events:
\begin{equation}
  e^* = \textsc{LocAgent}(\mathcal{M}_E \setminus \mathcal{S}_{\text{tried}},\; \mathcal{W},\; Q),
  \label{eq:loc_agent}
\end{equation}
where $\mathcal{S}_{\text{tried}}$ is the set of events already visited in the current query session.
By operating over progressively narrowed candidate sets, the agent ensures that each reasoning iteration explores a new temporal region, expanding the breadth of evidence gathering.

\noindent\textbf{Visual-Need Router.}
After an enriched text-only answer attempt fails to reach confidence, this agent decides whether pixel-level inspection of the selected event $e^*$ is likely to improve correctness:
\begin{equation}
  \textit{need\_visual} = \textsc{Router}(e^*,\; Q,\; r),
  \label{eq:router}
\end{equation}
where $r$ is the reason from the preceding answer attempt.
The router follows a perceptual-versus-temporal rubric.
It accepts (triggers frame extraction) when the question targets visual attributes such as appearance, on-screen text, spatial layout, or object disambiguation, i.e., evidence that cannot be reliably inferred from text alone.
It declines (skips frame extraction) when the question primarily concerns temporal ordering, scene transitions, or long-range narrative structure, where the textual memory already provides a suitable abstraction.
When declining, the router clears the visual working memory $\mathcal{W}$ and directs the loop to the next iteration without incurring any visual cost.
This role is unique to \method{abbr}: existing agent-based video systems~\cite{zuo2025videolucy,wang2025videotree,ma2025drvideo} do not gate visual access based on question type, and therefore either always or never use visual frames regardless of whether the question demands them.

Crucially, the router is \emph{not} a downstream perception module but an
explicit cost-control gate: it deliberately routes some questions away from
frames in order to keep the per-query frame budget tight.
This means the router may sacrifice marginal accuracy on attribute-centric
questions in exchange for a substantially lower average visual footprint, a
trade-off we view as a feature rather than a bug, because it exposes
per-query visual cost as a tunable hyperparameter rather than an emergent
byproduct of pipeline depth.
Applications that prioritise attribute fidelity over visual cost can simply
disable the router and always route to frames; we report both
configurations in our ablations (\cref{sec:experiment}).

\subsection{Story-First Reasoning Loop}
\label{sec:loop}

At query time, the system receives a question $Q$ and the pre-built memories $\mathcal{M}_E$ and $\mathcal{M}_C$.
It initializes the visual working memory $\mathcal{W} = \varnothing$, the explored set $\mathcal{S}_{\text{tried}} = \varnothing$, and a cumulative clip enrichment bank $\mathcal{B} = \varnothing$ that tracks which clip captions have been injected into which events.
The reasoning proceeds through the following steps, with a maximum budget of $T$ iterations.

\noindent\textbf{Step~1: Story-first answer.}
The Answering Agent receives the full Event Memory $\mathcal{M}_E$ and the empty $\mathcal{W}$, and attempts to answer $Q$.
If confident, the answer is returned immediately.
This early exit handles temporal and narrative questions that can be resolved from the story skeleton alone, requiring only a single LLM call for the entire query.
Importantly, CFD does not only route between textual and visual evidence.
Its story-first agentic loop also adaptively allocates the depth of reasoning
and evidence access according to whether the currently available memory is
sufficient.

\noindent\textbf{Step~2: Temporal localization.}
The Localization Agent selects the most relevant event $e^*$ from $\mathcal{M}_E \setminus \mathcal{S}_{\text{tried}}$, and $e^*$ is added to $\mathcal{S}_{\text{tried}}$.

\noindent\textbf{Step~3: Clip-enriched re-answer.}
All Clip Memory entries overlapping with $e^*$ are collected and injected as nested sub-descriptions within $e^*$'s event caption, and these enrichments are added to $\mathcal{B}$.
The narrative memory is then rebuilt: for every event $e_k$, if $\mathcal{B}$ contains clip enrichments for $e_k$, the clip captions are appended chronologically under a ``More details'' block within $e_k$'s text; the base event text is always restored first to prevent nesting accumulation.
The Answering Agent re-attempts with this enriched memory.
Because $\mathcal{B}$ accumulates across iterations, the narrative grows progressively richer as more events are explored, carrying forward all previously gathered local detail.
If confident, the answer is returned.

\noindent\textbf{Step~4: Visual-Need Router.}
If the answer is not confident, the Visual-Need Router evaluates whether pixel-level inspection of $e^*$ would help, following \cref{eq:router}.
If the router declines, $\mathcal{W}$ is cleared and the loop returns to Step~2 for the next iteration.

\noindent\textbf{Step~5: On-demand frame extraction.}
When the router accepts, keyframes are adaptively sampled from $e^*$.
The sampling rate is $f_{\text{ev}}$, capped at $N_{\text{fr}}$ frames per event; when the frame count exceeds $N_{\text{fr}}$, frames are redistributed uniformly across the event duration.
Extracted frames are pushed into the FIFO working memory $\mathcal{W}$ with capacity $N_{\text{wm}}$.

\noindent\textbf{Step~6: Multimodal re-answer.}
The Answering Agent receives the enriched narrative memory (with all cumulative clip injections) together with the visual evidence in $\mathcal{W}$, in a single multimodal prompt.
If confident, the answer is returned.

\noindent\textbf{Fallback.}
If $T$ iterations are exhausted without a confident answer, a must-answer fallback forces the agent to commit to its best answer using all accumulated textual and visual evidence.

Steps~2--6 repeat, and with each iteration the system's memory of the video becomes progressively richer: new events are explored (breadth), their clip details are permanently injected (depth), and visual evidence accumulates in $\mathcal{W}$ (grounding).
The full procedure is summarized as pseudocode in Algorithm~\ref{alg:cofd} of the appendix.

\section{Experiment}
\label{sec:experiment}

We evaluate \method{} on two long-video benchmarks: \textbf{Video-MME}~\cite{fu2025videomme}
under the standard \textit{without-subtitle} setting, and \textbf{InfiniBench}~\cite{ataallah2025infinibench}
on its four grounding-based skills (\textit{Chronological Understanding},
\textit{Character Actions Tracking}, \textit{Scene Transitions}, and \textit{Global Appearance}).
Full dataset descriptions are provided in Appendix~\ref{sec:supp_datasets}.

\subsection{Implementation details}
We use \texttt{Qwen3-VL-8B-Instruct}~\cite{bai2025qwen3vl} as the frozen edge-side Captioning Agent (MLLM) and \texttt{Qwen3-VL-32B-Instruct}~\cite{bai2025qwen3vl} as the cloud-side reasoning MLLM that instantiates the Answering Agent, Localization Agent, and Visual-Need Router.
Both models are served via vLLM~\cite{kwon2025vllm}.
\noindent\textbf{Offline indexing (edge).}
For the event memory, we apply TransNetV2~\cite{soucek2024transnet} (threshold $0.5$) for shot-boundary detection, with a minimum segment merging threshold $\tau_{\min} = 60$\,s.
Event-level captions use a structured prompt (scene overview, entity listing, chronological event flow, and anchor tags) at 1\,FPS.
For the clip memory, we use uniform clips of $\tau_c = 30$\,s with stride equal to $\tau_c$ (non-overlapping), captioned at 1\,FPS with a micro-action prompt (one action per sentence).
Both memory tracks are cached under a deterministic configuration fingerprint and reused across queries.
\noindent\textbf{Online reasoning (cloud).}
The maximum number of backtracking iterations is $T = 5$.
For on-demand frame extraction in Step~5, we use a sampling rate of $f_{\text{ev}} = 0.1$\,FPS (1 frame per 10\,s of event duration), capped at $N_{\text{fr}} = 8$ frames per event.
When the ideal frame count exceeds $N_{\text{fr}}$, frames are redistributed uniformly across the event span.
The FIFO visual working memory has capacity $N_{\text{wm}} = 16$.
The Visual-Need Router uses a perceptual-versus-temporal rubric and operates as a single LLM call with structured JSON output.
Detailed agent prompt designs are provided in Appendix~\ref{sec:impl}.

\begin{table*}[t]
\centering
\caption{Performance comparison on Video-MME~\cite{fu2025videomme}.
$\dagger$ denotes reproduced results under strictly identical setups using the authors' code.
\textit{\#Captions} denotes the average temporal span covered by each caption.
\textit{Re-caption?} indicates whether a method revisits video frames at query time to (re)caption or refine its textual memory.
\textit{\#Frames} counts the number of frames fed to the answering model at the final  inference stage.}
\resizebox{\linewidth}{!}{%
\begin{tabular}{lccccccc}
\hline
\multirow{2}[4]{*}{\textbf{Method}} & \multirow{2}[4]{*}{\textbf{\#Captions}} & \multirow{2}[4]{*}{\textbf{Re-caption?}} & \multirow{2}[4]{*}{\textbf{\#Frames}} & \multicolumn{4}{c}{\textbf{Video-MME}} \bigstrut\\
\cline{5-8}      &       &       &       & \textbf{short} & \textbf{medium} & \textbf{long} & \textbf{overall} \bigstrut\\
\hline
\hline
\textit{Leading Open-source MLLMs} &       &       &       &       &       &       &  \bigstrut[t]\\
VideoChat2-7B~\cite{li2024mvbench} & -     & -     & 16    & 48.3  & 37.0  & 33.2  & 39.5 \\
LongVA-7B~\cite{zhang2024longva} & -     & -     & 128   & 61.1  & 50.4  & 46.2  & 52.6 \\
Kangaroo-7B~\cite{liu2024kangaroo} & -     & -     & 64    & 66.1  & 55.3  & 46.6  & 56.0 \\
Video-CCAM-14B~\cite{fei2024video} & -     & -     & 96    & 62.2  & 50.6  & 46.7  & 53.2 \\
VideoXL-7B~\cite{shu2025videoxl} & -     & -     & 128   & 64.0  & 53.2  & 49.2  & 55.5 \\
Dispider-7B~\cite{qian2025dispider} & -     & -     & 1 fps & -     & -     & -     & 57.2 \\
VideoChat-Online-4B~\cite{huang2025online} & -     & -     & 2 fps & -     & -     & 47.1  & 54.4 \\
TimeChat-Online-7B~\cite{yao2025timechatonline} & -     & -     & 1 fps & -     & -     & 48.4  & 62.4 \\
Qwen3-VL-32B~\cite{bai2025qwen3vl} & -     & -     & 768   & 82.4  & 76.3  & 69.0  & 75.9 \bigstrut[b]\\
\hline
\textit{Agent-based Systems} &       &       &       &       &       &       &  \bigstrut[t]\\
VideoAgent~\cite{fan2024videoagent} & 2s / caption & \cmark & -     & -     & -     & 46.4  & - \\
VideoTree~\cite{wang2025videotree} & 8s / caption & \cmark & -     & 67.8  & 59.9  & 54.2  & 60.6 \\
DrVideo~\cite{ma2025drvideo} & 5s / caption & \cmark & -     & -     & -     & 51.7  & - \\
MemVid~\cite{yuan2025memory} & -     & \xmark & 1 fps & \textbf{73.9} & 63.1  & 55.0  & 64.0 \\
VideoLucy~\cite{zuo2025videolucy}$^\dagger$ & < 30s / caption & \cmark & -     & 73.2  & 64.7  & 56.2  & 64.7 \bigstrut[b]\\
\hline
\rowcolor[rgb]{ .91,  .91,  .91} \method{} (Ours) & \textbf{233.1s / caption} & \xmark & \textbf{5.8} & 72.2  & \textbf{66.6} & \textbf{63.6} & \textbf{67.5} \bigstrut\\
\hline
\end{tabular}%

}
\label{tab:videomme}
\end{table*}

\begin{table*}[t]
\centering
\caption{Performance comparison on InfiniBench~\cite{ataallah2025infinibench}.
\textit{\#Captions} denotes the average temporal span covered by each caption.
\textit{\#Frames} counts the number of frames fed to the answering model at the final  inference stage.
\textit{Chronological Understanding} evaluates ordering events across the full video;
\textit{Character Actions Tracking} requires aggregating and sequencing a character's actions;
\textit{Scene Transitions} tests recognizing and ordering scene/location shifts;
\textit{Global Appearance} tracks long-term appearance changes of a character (e.g., outfit changes).}
\resizebox{\linewidth}{!}{%
\begin{tabular}{lcccccc}
\hline
\multirow{2}[2]{*}{\textbf{Method}} & \multirow{2}[2]{*}{\textbf{\#Captions}} & \multirow{2}[2]{*}{\textbf{\#Frames}} & \textbf{Chronological } & \textbf{Global } & \textbf{Scene } & \textbf{Character } \bigstrut[t]\\
      &       &       & \textbf{Understanding } & \textbf{Appearance} & \textbf{Transitions} & \textbf{Actions} \bigstrut[b]\\
\hline
\hline
LLaVA-OneVision~\cite{li2024llavaonevision} & -     & 128   & 43.91 & 37.21 & 25.40 & 20.11 \bigstrut[t]\\
InternVL2.5~\cite{chen2024internvl25} & -     & 128   & 42.16 & 34.88 & 20.63 & 21.60 \\
Qwen2-VL~\cite{wang2024qwen2} & -     & 768   & 48.41 & 31.01 & 23.81 & 32.59 \\
Qwen2.5-VL~\cite{bai2025qwen25vl} & -     & 768   & 27.17 & 37.98 & 22.22 & 22.53 \\
InternVL3~\cite{zhu2025internvl3} & -     & 128   & 36.29 & 33.33 & 28.57 & 23.46 \\
Qwen3-VL~\cite{bai2025qwen3vl} & -     & 768   & 48.44 & 70.54 & 53.97 & 67.04 \bigstrut[b]\\
\hline
\rowcolor[rgb]{ .91,  .91,  .91} \method{} (Ours) & 206s / caption & 14.8  & 55.10 & 58.90 & 52.40 & 56.20 \bigstrut\\
\hline
\end{tabular}%

}
\label{tab:infinibench}
\end{table*}

\begin{table*}[t]
\footnotesize
\centering
\setlength{\tabcolsep}{3pt}
\renewcommand{\arraystretch}{0.85}
\caption{Component Study. (\textsuperscript{*}) denotes that all clip captions in one video are used for the ablation experiment.}
\resizebox{\linewidth}{!}{%
\begin{tabular}{ccccccccccccc}
\hline
\multirow{3}[4]{*}{\textbf{Event}} & \multirow{3}[4]{*}{\textbf{Clip}} & \multicolumn{1}{c}{\multirow{3}[4]{*}{\textbf{Reasoning \newline{}Loop}}} & \multirow{3}[4]{*}{\textbf{Frames}} & \multirow{3}[4]{*}{\textbf{Router}} & \multicolumn{4}{c}{\textbf{Video-MME}} & \multicolumn{4}{c}{\textbf{InfiniBench}} \bigstrut\\
\cline{6-13}      &       &       &       &       & \multirow{2}[2]{*}{\textbf{short}} & \multirow{2}[2]{*}{\textbf{medium}} & \multirow{2}[2]{*}{\textbf{long}} & \multirow{2}[2]{*}{\textbf{overall}} & \textbf{Chronological } & \textbf{Global } & \textbf{Scene } & \textbf{Character } \bigstrut[t]\\
      &       &       &       &       &       &       &       &       & \textbf{Understanding } & \textbf{Appearance} & \textbf{Transitions} & \textbf{Actions} \bigstrut[b]\\
\hline
\hline
\cmark &       &       &       &       & 62.0  & 53.7  & 56.3  & 57.3  & 52.30 & 9.30  & 39.70 & 16.90 \bigstrut[t]\\
      & \cmark\textsuperscript{*}  &       &       &       & 68.2  & 66.4  & 62.0  & 65.5  & 55.70 & 38.00 & 65.10 & 43.90 \\
\cmark & \cmark & \cmark &       &       & 68.8  & 64.0  & 63.0  & 65.3  & 55.10 & 45.00 & 52.40 & 53.40 \\
\cmark & \cmark & \cmark & \cmark &       & 72.3  & 66.7  & 62.6  & 67.2  & 51.70 & 62.80 & 58.70 & 65.90 \bigstrut[b]\\
\hline
\cmark & \cmark & \cmark & \cmark & \cmark & 72.2  & 66.6  & 63.6  & 67.5  & 55.10 & 58.90 & 52.40 & 56.20 \bigstrut\\
\hline
\end{tabular}%

}
\label{tab:component}
\end{table*}

\begin{table*}[t]
\footnotesize
\begin{minipage}[t]{0.55\linewidth}
\centering
\setlength{\tabcolsep}{3pt}
\renewcommand{\arraystretch}{0.85}
\caption{Comparison across different captioner model sizes.}
\resizebox{0.8\linewidth}{!}{%
\begin{tabular}{lccccc}
\hline
Caption Model & Size  & short & medium & long  & overall \bigstrut\\
\hline
\hline
\multirow{3}[2]{*}{Qwen3-VL} & 2B    & 67.7  & 61.9  & 58.1  & 62.6 \bigstrut[t]\\
      & 8B    & 72.2  & 66.6  & 63.6  & 67.5 \\
      & 32B   & 72.4  & 67.1  & 61.0  & 66.8 \bigstrut[b]\\
\hline
\end{tabular}%

}
\label{tab:captioner_size}
\end{minipage}
\hfill
\begin{minipage}[t]{0.43\linewidth}
\centering
\setlength{\tabcolsep}{3pt}
\renewcommand{\arraystretch}{0.85}
\caption{Comparison across different budget settings.}
\resizebox{0.8\linewidth}{!}{%
\begin{tabular}{cccccc}
\hline
$N_{\text{fr}}$ & $N_{\text{wm}}$ & short & medium & long  & overall \bigstrut\\
\hline
\hline
2     & 4     & 71.8  & 66.3  & 63.6  & 67.2 \bigstrut[t]\\
8     & 16    & 72.2  & 66.6  & 63.6  & 67.5 \\
32    & 64    & 72.1  & 66.6  & 64.2  & 67.6 \bigstrut[b]\\
\hline
\end{tabular}%

}
\label{tab:FIFO}
\end{minipage}
\vspace{-2mm}
\end{table*}

\subsection{Comparison with other methods}
Table~\ref{tab:videomme} reports results on Video-MME.
Open-sourced MLLMs typically feed hundreds to near-thousands of densely
sampled frames; for instance, Qwen3-VL-32B reaches 75.9 overall at 768 frames
per question.
\method{} achieves \textbf{67.5} overall with only \textbf{5.8 frames per
question} on average, more than two orders of magnitude fewer than
dense baselines.
This asymmetry suggests that competitive temporal reasoning does not
require saturating the context window with visual tokens, provided the
right frames are retrieved at the right time.
Among agent-based systems, VideoAgent~\cite{fan2024videoagent},
VideoTree~\cite{wang2025videotree}, DrVideo~\cite{ma2025drvideo}, and
VideoLucy~\cite{zuo2025videolucy} densely caption the full video during
preprocessing and further revisit raw frames at query time to refine their
textual memory, while MemVid~\cite{yuan2025memory} avoids re-captioning but
compensates with dense 1\,fps uniform sampling (64.0 overall, 73.9 on short
videos).
\method{} matches or exceeds all reported agent-based baselines
(VideoLucy 64.7, MemVid 64.0) while using roughly an order of magnitude
fewer frames per question.
We do not aim to match dense-inference upper bounds such as Qwen3-VL-32B at
768 frames; we target the accuracy-efficiency frontier under bounded visual
budgets.
The gains over agent-based methods are most pronounced on medium and long
videos (66.6 and 63.6), where re-captioning and dense sampling suffer most
from context saturation and redundant visual processing.

Table~\ref{tab:infinibench} reports results on the four grounding-based
skills of InfiniBench.
\method{} achieves the best Chronological Understanding score (55.10),
exceeding the 768-frame dense baselines Qwen3-VL (48.44) and Qwen2-VL
(48.41), and remains competitive on Scene Transitions (52.40 \vs\ 53.97 for
Qwen3-VL), at only \textbf{14.8 frames per question} on average.
These two skills are most aligned with our design principle: ordering events
and recognising scene-level shifts are temporally structured tasks for
which the event-level narrative skeleton provides a compact discrete record.
On Global Appearance and Character Actions, \method{} (58.90 / 56.20) is
ahead of agent-based baselines but below the dense Qwen3-VL, consistent
with the visual-textual duality identified in Section~\ref{sec:intro}:
fine-grained attribute binding benefits from dense pixel access, and a
tight on-demand budget does not fully substitute for exhaustive sampling.
Even so, within the agent-based regime the Visual-Need Router still
recovers a substantial fraction of the appearance-grounding evidence under
a tight frame budget.

\subsection{Ablation study}

\noindent\textbf{Component Analysis.}
Table~\ref{tab:component} ablates each component.
Event memory alone is too sparse (57.3 overall, Global Appearance 9.30).
Dense clip memory alone (Row~2) recovers most of the gap (65.5 overall,
Chronological Understanding 55.70) but Global Appearance remains limited
(38.00), confirming that text alone cannot substitute for visual evidence.
Adding the reasoning loop over the dual-track memory without any frames
(Row~3) achieves comparable overall performance (65.3), corroborating
that language memories are a strong carrier for temporal structure while
attribute perception remains the bottleneck of text-only reasoning.
Enabling on-demand frames without the router (Row~4) brings sharp gains on
appearance-centric tasks (Global Appearance 62.80, Character Actions 65.90)
at the cost of Chronological Understanding (51.70) and long-video accuracy
(62.6), reflecting that indiscriminate frame injection dilutes the
narrative context on temporally structured questions.
The full model with the Visual-Need Router (Row~5) suppresses frames on
those questions, recovering Chronological Understanding to 55.10 and
long-video accuracy to 63.6 while keeping the average frame count low,
achieving a more balanced accuracy-efficiency trade-off than always-on
frame retrieval.
This confirms that text and frames are complementary: text excels at
long-horizon temporal structure, while frames remain decisive for
fine-grained attribute perception, and routing between them captures the
best of both regimes.

\noindent\textbf{Offline captioner model size.}
Table~\ref{tab:captioner_size} sweeps three Qwen3-VL captioner sizes used
during offline indexing.
Scaling \texttt{2B}$\rightarrow$\texttt{8B} brings a substantial gain
(62.6$\rightarrow$67.5 overall); scaling further to \texttt{32B} marginally
helps short and medium videos but degrades long-video accuracy
(61.0 \vs\ 63.6).
We attribute this to verbosity accumulation: larger captioners produce
longer, more detailed descriptions per segment, which compound across the
many events of a long video and inflate the total narrative context beyond
what the reasoning agent can attend to within a fixed inference budget.
We therefore adopt \texttt{Qwen3-VL-8B} as the default, balancing indexing
cost, caption conciseness, and downstream accuracy.
The \texttt{Qwen3-VL-2B} captioner reaches 62.6 overall and provides a lower-resource alternative for more constrained offline indexing tiers.

\noindent\textbf{Budget sensitivity.}
Table~\ref{tab:FIFO} sweeps the per-event frame cap $N_{\text{fr}}$ and
FIFO capacity $N_{\text{wm}}$ over an 8$\times$ range.
Overall accuracy varies by only 0.4\%, indicating that a few targeted
frames suffice; the middle setting
($N_{\text{fr}}{=}8,\,N_{\text{wm}}{=}16$) is within 0.1\% of the largest
budget at one-quarter the visual cost.
Long videos benefit most (63.6$\to$64.2), consistent with sparser event
memory leaving more room for visual verification.
These results support the core premise of \method{}: a small, bounded set
of on-demand frames closes most of the gap that narrative memory alone
cannot cover.

\begin{table}[t]
\centering
\caption{Visual-budget sweep on Video-MME.}
\resizebox{0.72\linewidth}{!}{%
\begin{tabular}{cccc}
\hline
$N_{\mathrm{fr}}$ & $N_{\mathrm{wm}}$ & Frames/query & Accuracy \\
\hline
\hline
2 & 4 & 1.83 & 67.2 \\
8 & 16 & 5.76 & 67.5 \\
32 & 64 & 8.93 & 67.6 \\
\hline
\end{tabular}

}
\label{tab:budget_detail}
\end{table}

\noindent\textbf{Frame reduction and routing.}
Even when the Router is disabled and every localized event is routed to frames,
the caption-once memory, temporal localization, iterative backtracking,
bounded frame extraction, and fixed-capacity visual working memory reduce
online frame usage from 768 to 6.63 frames/query, corresponding to
\textbf{99.1\% fewer frames}, or approximately a \textbf{116$\times$}
reduction. Within this already sparse pipeline, the Visual-Need Router further
reduces frame usage from 6.63 to 5.76 frames/query, a further
\textbf{13.1\% reduction}, while improving accuracy from 67.2 to 67.5
(\textbf{+0.3 percentage points}). Frame savings do not imply lower wall-clock
latency in every configuration, because the prompt-based Router introduces an
additional LLM call.
The matched budget sweep reports the corresponding online frame counts in Table~\ref{tab:budget_detail}.
Increasing the budget from 2/4 to 8/16 improves accuracy by 0.3 points,
while further increasing it to 32/64 yields only another 0.1 point despite a
substantially larger frame footprint. We therefore treat 8/16 as an
accuracy--cost operating point rather than exhaustive visual coverage.

\begin{table}[t]
\centering
\caption{Controlled routing comparison on Video-MME. Each cell is (accuracy
 / online frames per query).}
\setlength{\tabcolsep}{3.5pt}
\resizebox{\linewidth}{!}{%
\begin{tabular}{lcccc}
\hline
Policy & Short & Medium & Long & Average \\
\hline
\hline
No frames & 68.8 / 0.00 & 64.0 / 0.00 & 63.0 / 0.00 & 65.3 / 0.00 \\
Visual-Need Router & 72.2 / 2.38 & 66.6 / 5.81 & 63.6 / 9.09 & 67.5 / 5.76 \\
Always frames & 72.3 / 2.54 & 66.7 / 6.57 & 62.6 / 10.78 & 67.2 / 6.63 \\
\hline
\end{tabular}

}
\label{tab:routing_policy}
\vspace{-5mm}
\end{table}

\noindent\textbf{Routing policy.}
Table~\ref{tab:routing_policy} reports the controlled comparison among no-frame,
Visual-Need Router, and always-frame inference. The Router improves average
accuracy by \textbf{2.2 points} over no-frame reasoning and slightly exceeds
always-frame inference (\textbf{67.5 vs. 67.2}) while reducing online frames by
\textbf{13.1\%} on average. The full policy table and outcome-oracle analysis
are provided in Appendix~\ref{sec:supp_ablation}.
Table~\ref{tab:router_trigger} in Appendix~\ref{sec:supp_ablation} reports
trigger statistics from the same paired diagnostic rerun used to construct the
outcome oracle. The Router retrieves frames for \textbf{46.2\%} of questions
and therefore leaves \textbf{53.8\%} entirely in language space, while
capturing 127 of 133 (\textbf{95.5\%}) cases in which frames empirically
improve the answer. These cases are outcome-based diagnostics, not human
routing annotations.

\begin{figure}[t]
\centering
\includegraphics[width=0.7\linewidth]{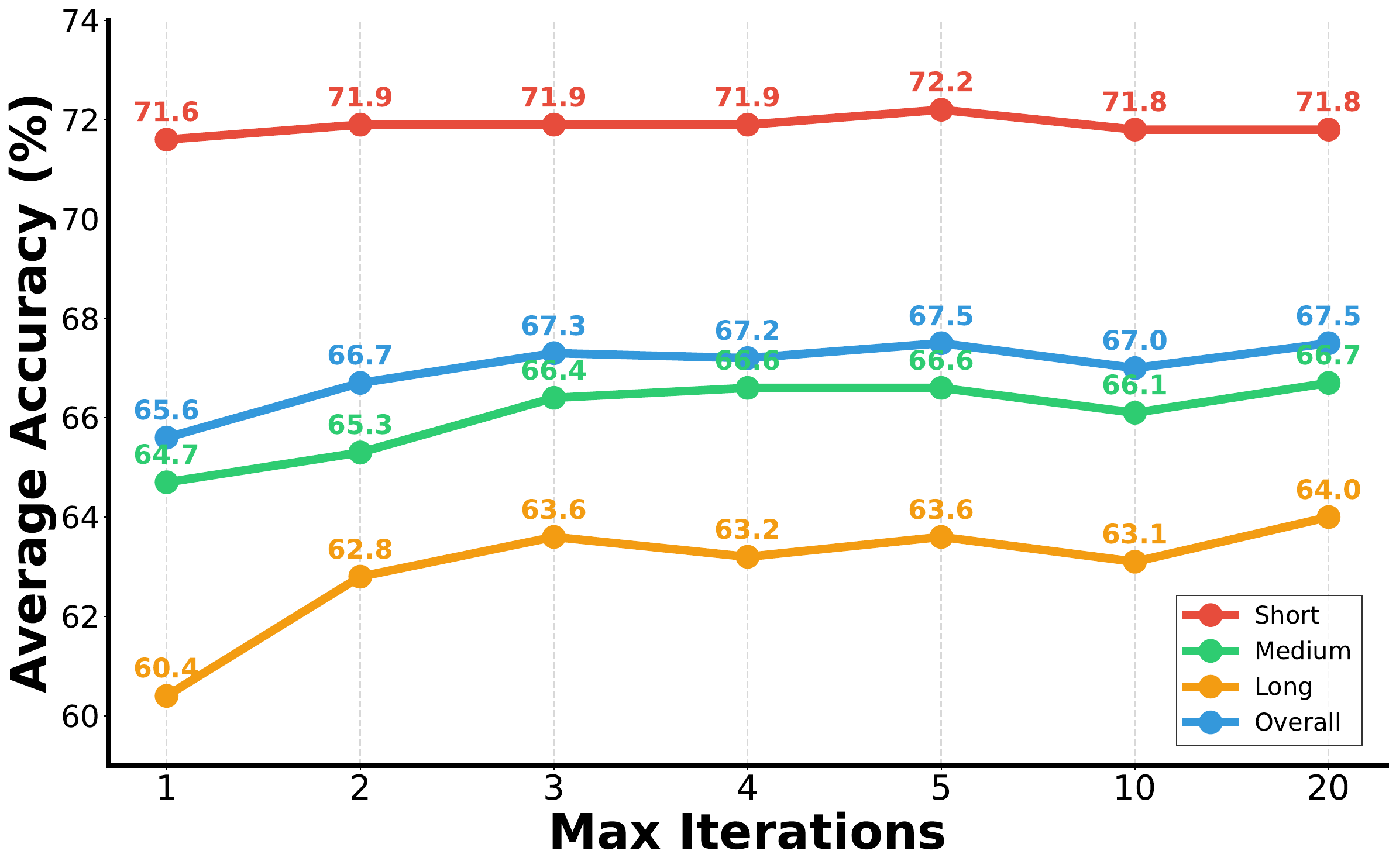}
\caption{Effect of the maximum iteration budget $T$ on Video-MME accuracy
across short, medium, long, and overall splits.}
\label{fig:max_iter}
\vspace{-2mm}
\end{figure}

\noindent\textbf{Iteration budget.}
Figure~\ref{fig:max_iter} shows performance as a function of the maximum
iteration budget $T$. Short-video accuracy saturates rapidly, reaching 71.9 at
$T{=}2$, while medium-video performance improves until $T{=}4$ (66.6) and then
plateaus. Long-video accuracy continues improving from 60.4 at $T{=}1$ to 64.0
at $T{=}20$, indicating that longer videos benefit from additional retrieval
iterations. We adopt $T{=}5$ as the default, which achieves the best overall
score of 67.5 and provides a favorable trade-off across duration splits.

\begin{table}[t]
\centering
\caption{Clip-caption activation on Video-MME. Counts are captions per query,
not seconds per caption or captioning runtime.}
\resizebox{\linewidth}{!}{%
\begin{tabular}{lcccc}
\hline
Method & Short & Medium & Long & Average \\
\hline
\hline
All 30-s clip captions & 3.10 & 17.69 & 82.66 & 34.49 \\
\method{}-activated captions & 1.52 & 9.29 & 24.05 & 11.62 \\
Caption reduction & 51.2\% & 47.5\% & 70.9\% & 66.3\% \\
\hline
\end{tabular}

}
\label{tab:caption_reduction}
\vspace{-2mm}
\end{table}

\noindent\textbf{Clip-caption activation.}
As shown in Table~\ref{tab:caption_reduction}, CFD uses \textbf{66.3\% fewer
clip captions on average} while outperforming the all-clip baseline overall.
Event Memory uses TransNetV2 boundaries, whereas Clip Memory is generated
uniformly and independently of event segmentation.
The outcome oracle uses frames only when the always-frame policy corrects an
error made by the no-frame policy; it is an empirical upper bound rather than
semantic ground-truth annotation. Its 133 frame-beneficial cases correspond
to a \textbf{4.9\%} trigger rate over all 2,700 questions, distinct from the
Router's 46.2\% YES rate. The oracle obtains 69.5\% accuracy with 0.64
frames/query, compared with the Router's 67.5 accuracy and 5.76 frames/query.

\section{Conclusion}
\label{sec:conclusion}

We presented \method{abbr}, a budget-aware edge-cloud framework that 
indexes video into reusable narrative memory once and retrieves sparse 
visual frames only when language alone cannot ground the answer.
Experiments across long-video benchmarks confirm that this
caption-once, frames-on-demand design achieves competitive accuracy while 
substantially reducing online visual cost.

\clearpage
\section*{Limitations}
\noindent\textbf{Caption quality and segmentation sensitivity.}
Our framework is fundamentally bottlenecked by the quality of the
offline captioning pass. If question-critical visual details are not
captured in the question-agnostic memories, purely text-based
backtracking cannot recover them afterwards. Sparse visual
backtracking partially alleviates this issue, but its effectiveness
still depends on whether the system can correctly diagnose why
text-only reasoning failed and route the query to the appropriate
event. Moreover, even after the correct event is localized, the bounded
keyframe extraction stage may still miss highly transient cues, small
objects, or brief textual overlays that fall outside the sampled
frames. It may also miss subtle motion. Failure of both textual
localization and sparse visual retrieval therefore remains possible.
The framework is also sensitive to the quality of upstream
event segmentation: overly coarse segmentation can bury useful evidence
inside broad summaries, whereas overly fine segmentation may fragment
semantically coherent episodes and weaken later localization.
On the modeling side, both the captioner and the Visual-Need Router
currently operate in a generic zero-shot setting, and would likely
benefit from domain-specific supervision or lightweight fine-tuning to
improve structured caption fidelity and routing accuracy. Finally, our
edge-cloud setup is validated under simulated separation rather than
physical deployment, so the practical impact of device-side compute
constraints, communication overhead, and network latency variability
remains to be systematically characterized on representative edge
hardware.

\noindent\textbf{Router granularity.}
The current Visual-Need Router operates as a binary, zero-shot
classifier over a perceptual-versus-temporal rubric. While this coarse
dichotomy is sufficient to demonstrate that question-type-conditioned
visual gating yields meaningful cost savings, it inevitably mis-routes
queries that fall on the boundary (e.g., a chronological question whose
answer hinges on a specific visual change). A continuous, calibrated
confidence score, possibly trained from a small held-out routing set
or distilled from oracle traces, would likely improve per-skill
balance, particularly on InfiniBench's attribute-heavy skills where the
current router under-allocates frames relative to the no-router
configuration (Row~4 of Table~\ref{tab:component}).

\noindent\textbf{Egocentric and content-dependent segmentation.}
Our event memory relies on TransNetV2 shot-boundary detection, which is
effective for edited video (TV shows, documentaries, lectures) but is
not designed for continuous egocentric or wearable recordings where
hard cuts are rare. Adapting \method{} to egocentric streams would
require replacing shot-boundary segmentation with an audio-visual event
detector (e.g., EgoTrigger~\cite{paruchuri2025egotrigger}) or an
LLM-prompted scene-change estimator over coarse-frame thumbnails; both
are orthogonal to our framework and represent natural follow-up work.

\noindent\textbf{Definition of ``edge''.}
We use ``edge'' to denote a separation of inference tiers (offline
indexing vs.\ online cloud reasoning) rather than a claim that the
captioner runs on mobile-class hardware. Our default captioner
(Qwen3-VL-8B) targets a modest server-class device or a powerful
desktop with a single GPU; deploying onto smartphones or smart glasses
would require additional model compression (e.g., Qwen3-VL-2B with
quantisation), which we leave to future work alongside end-to-end
latency and energy measurements on representative hardware.

\bibliography{main}

\clearpage
\twocolumn
\appendix
This appendix complements the main paper with additional analyses,
organized as follows.

\textbf{Appendix~\ref{sec:supp_algorithm} (Reasoning Loop Algorithm)} presents
the pseudocode of the story-first reasoning loop with on-demand frame
retrieval that orchestrates the four agent roles described in the main paper.

\textbf{Appendix~\ref{sec:supp_datasets} (Datasets)} provides full
descriptions of the Video-MME and InfiniBench benchmarks used in the main
experiments, together with the evaluation protocol and the rationale for
focusing on InfiniBench's grounding-based skills.

\textbf{Appendix~\ref{sec:related_work} (Related Work)} positions \method{}
with respect to prior work on long video understanding with MLLMs, agentic
video understanding, and video understanding on edge devices.

\textbf{Appendix~\ref{sec:supp_lvbench} (Results on LVBench)} extends the
main evaluation to LVBench, a challenging hour-scale benchmark,
demonstrating the generalizability of \method{}.

\textbf{Appendix~\ref{sec:supp_ablation} (Further Ablation Study)} studies
cloud reasoning model size, cross-family backbone generalization, component
and budget sensitivity, inference efficiency, practical cost accounting, and
Visual-Need Router reliability. It also presents qualitative visualizations of
representative inference trajectories illustrating the story-first,
evidence-on-demand behavior.

\textbf{Appendix~\ref{sec:impl} (More Implementation Details)} supplies the
full agent prompts used in our system.

\section{Reasoning Loop Algorithm}
\label{sec:supp_algorithm}

For completeness, Algorithm~\ref{alg:cofd} presents the pseudocode of the
story-first reasoning loop with on-demand frame retrieval described in
Section~2.3 of the main paper.
The procedure operates over the dual-track narrative memory $(\mathcal{M}_E,
\mathcal{M}_C)$ and the visual working memory $\mathcal{W}$, invoking the
Answering, Localization, and Visual-Need Router agents under a bounded
iteration budget $T$ and a fixed frame budget governed by $N_{\text{fr}}$
and the FIFO capacity $N_{\text{wm}}$.

\begin{algorithm*}[h]
\caption{Story-First Reasoning with On-Demand Frames}
\label{alg:cofd}
\begin{algorithmic}[1]
\Require Video $V$, question $Q$, memories $\mathcal{M}_E$, $\mathcal{M}_C$, agents $\textsc{AnsAgent}$, $\textsc{LocAgent}$, $\textsc{Router}$, budget $T$, FIFO capacity $N_{\text{wm}}$, max frames $N_{\text{fr}}$, sampling rate $f_{\text{ev}}$.

\State $\mathcal{W} \gets \varnothing$;\; $\mathcal{S}_{\text{tried}} \gets \varnothing$;\; $\mathcal{B} \gets \varnothing$ \Comment{working memory, explored set, clip bank}

\Comment{\textbf{Step 1}: Story-first answer}
\State $(a, \textit{conf}, r) \gets \textsc{AnsAgent}(\mathcal{M}_E, \mathcal{W}, Q)$
\If{$\textit{conf}$} \Return $a$ \EndIf

\For{$i = 1$ \textbf{to} $T$}

  \Comment{\textbf{Step 2}: Temporal localization}
  \State $e^* \gets \textsc{LocAgent}(\mathcal{M}_E \setminus \mathcal{S}_{\text{tried}},\, \mathcal{W},\, Q)$
  \State $\mathcal{S}_{\text{tried}} \gets \mathcal{S}_{\text{tried}} \cup \{e^*\}$

  \Comment{\textbf{Step 3}: Clip-enriched re-answer}
  \State $\mathcal{B} \gets \mathcal{B} \cup \{(e^*,\, \{m_j^C : c_j \cap e^* \neq \varnothing\})\}$ \Comment{accumulate clip enrichments}
  \State $\mathcal{M} \gets \textsc{Rebuild}(\mathcal{M}_E,\, \mathcal{B})$ \Comment{inject all accumulated clips}
  \State $(a, \textit{conf}, r) \gets \textsc{AnsAgent}(\mathcal{M},\, \mathcal{W},\, Q)$
  \If{$\textit{conf}$} \Return $a$ \EndIf

  \Comment{\textbf{Step 4}: Visual-Need routing}
  \State $\textit{need\_visual} \gets \textsc{Router}(e^*,\, Q,\, r)$
  \If{$\neg\,\textit{need\_visual}$}
    \State $\mathcal{W} \gets \varnothing$;\; \textbf{continue} \Comment{skip frames, next iteration}
  \EndIf

  \Comment{\textbf{Step 5}: On-demand frame extraction}
  \State $\mathcal{F} \gets \textsc{AdaptiveSample}(V,\, e^*,\, f_{\text{ev}},\, N_{\text{fr}})$
  \State $\mathcal{W} \gets \textsc{FIFO\_Push}(\mathcal{W},\, \mathcal{F},\, N_{\text{wm}})$

  \Comment{\textbf{Step 6}: Multimodal re-answer}
  \State $(a, \textit{conf}, r) \gets \textsc{AnsAgent}(\mathcal{M},\, \mathcal{W},\, Q)$
  \If{$\textit{conf}$} \Return $a$ \EndIf

\EndFor

\State \Return $\textsc{MustAnswer}(\mathcal{M},\, \mathcal{W},\, Q)$ \Comment{Fallback}
\end{algorithmic}
\end{algorithm*}

\section{Datasets}
\label{sec:supp_datasets}

\noindent\textbf{Video-MME.}
Video-MME~\cite{fu2025videomme} is a comprehensive multimodal evaluation benchmark for video understanding, containing 900 manually curated videos and 2,700 expert-annotated multiple-choice questions (3 per video).
The benchmark covers 6 primary visual domains with 30 fine-grained subcategories and spans a wide range of durations: short ($<$2 min, avg.\ 82.5s), medium (4--15 min, avg.\ 562.7s), and long (30--60 min, avg.\ 2,385.5s).
Question types encompass temporal perception, spatial reasoning, action recognition, object recognition, and information synopsis, among others.
All experiments follow the standard \textit{without-subtitle} setting, and we report accuracy on each duration split as well as the overall average.

\noindent\textbf{InfiniBench.}
InfiniBench~\cite{ataallah2025infinibench} is a large-scale benchmark targeting long-form video understanding in movies and TV episodes, featuring an average video duration of 53 minutes and over 87.7K question-answer pairs.
We evaluate our system exclusively on its four \emph{grounding-based} skills, which assess a model's ability to retrieve, order, and structure video content without requiring causal inference:
\textit{Chronological Understanding} measures the ability to correctly sequence a set of events across the full video. \textit{Character Actions Tracking} requires grouping and ordering all actions performed by a specific character over the video duration. \textit{Scene Transitions} tests recognition and sequential ordering of scene-level location shifts.
\textit{Global Appearance} evaluates long-term tracking of changes in a character's visual appearance (\eg, the sequence of outfit changes throughout the video).
All four skills are posed as multiple-choice questions and evaluated with standard classification accuracy.
We focus on these skills because they directly probe temporally-extended visual localization, the core capability our system is designed to address, without conflating performance with the higher-level causal and narrative inference required by the reasoning-based skills.

\section{Related Work}
\label{sec:related_work}

\noindent\textbf{Long video understanding with MLLMs.}
Benchmarks from EgoSchema~\cite{mangalam2023egoschema} to hour-scale suites such as Video-MME~\cite{fu2025videomme}, MLVU~\cite{zhou2025mlvu}, LVBench~\cite{wang2025lvbench}, and LongVideoBench~\cite{wu2024longvideobench} consistently show that performance degrades with duration, motivating work along two axes.
\emph{Visual compression} methods reduce token count so that more frames fit within an MLLM's context: Video-XL~\cite{shu2025videoxl} condenses KV states across intervals, LongVU~\cite{shen2024longvu} filters redundant frames via DINOv2 similarity, and VideoChat-Flash~\cite{li2024videochatflash} applies hierarchical token merging~\cite{bolya2022tome}.
\emph{Textual translation} methods instead convert video into language surrogates: LLoVi~\cite{zhang2024simple} chains dense captioning with LLM summarization, LangRepo~\cite{kahatapitiya2025language} maintains a prunable language repository, and TOPA~\cite{li2024topa} demonstrates temporal reasoning from synthetic textual videos alone.
Text compactly encodes temporal topology but systematically loses fine-grained visual attributes~\cite{yang2025captionqa}, motivating our factorization of language narrative and on-demand visual evidence.
The observation that captions can carry much of the long-range temporal
reasoning load is already implicit in LLoVi (EgoSchema results) and made
explicit in TOPA; we therefore treat this visual-textual duality as
established prior context rather than a primary contribution, and focus
instead on how to systematically \emph{gate} visual access via a
query-conditioned router so that the duality is exploited under explicit
cost control rather than left to emergent pipeline behaviour.

\noindent\textbf{Agentic video understanding.}
Rather than processing all frames at once, agent-based systems treat videos as searchable environments.
Early examples include VideoAgent~\cite{fan2024videoagent} (iterative planning and retrieval), DrVideo~\cite{ma2025drvideo} (document-augmented agent loops), and VideoTree~\cite{wang2025videotree} (query-adaptive keyframe trees).
Recent work introduces richer memory and search: VideoLucy~\cite{zuo2025videolucy} proposes hierarchical memory with iterative backtracking; VideoARM~\cite{yin2025videoarm}, DVD~\cite{zhang2025deep}, HAVEN~\cite{lai2025haven}, EGAgent~\cite{rege2026agentic}, and EventMemAgent~\cite{wen2026eventmemagent} further explore dynamic multimodal memory, multi-granular toolsets, entity-aware indexing, and online memory with agentic RL.
Ours shares the evidence-gathering paradigm but \emph{factorises} the representation into a reusable narrative index plus on-demand visual evidence, and explicitly optimizes for edge-cloud cost rather than accuracy alone.
Several recent agentic systems (DVD~\cite{zhang2025deep},
EGAgent~\cite{rege2026agentic}, EventMemAgent~\cite{wen2026eventmemagent})
appeared concurrently with this work; we do not include head-to-head
numbers because their public implementations were not yet available at the
time of submission, and we leave systematic cross-comparison to follow-up
work.

\noindent\textbf{Video understanding on edge devices.}
Collaborative intelligence~\cite{kang2017neurosurgeon} and deep feature compression~\cite{choi2018deep} study DNN partitioning and intermediate-feature codecs for split inference but target single-inference offloading.
EgoTrigger~\cite{paruchuri2025egotrigger} gates camera activation on smart glasses via audio cues, reducing capture by 54\% while preserving QA accuracy.
Our framework operates at a higher semantic level: instead of gating the camera, we gate the \emph{representation} sent to the cloud, transmitting compact narrative memory by default and reserving frame uploads for moments requiring visual attribution.

\section{Results on LVBench}
\label{sec:supp_lvbench}
In this section, we conduct more comprehensive comparisons on the LVBench~\cite{wang2025lvbench}.
LVBench~\cite{wang2025lvbench} is a benchmark specifically designed for long video understanding. 
It contains 103 manually curated long-form YouTube videos spanning 117 hours in total, with an average duration of 4{,}101 seconds per video, making it substantially longer than most existing video understanding benchmarks. 
The videos cover six diverse domains, including sports, documentary, event record, lifestyle, TV shows, and cartoons, and are paired with 1,549 human-annotated question-answer instances. 
To comprehensively evaluate long-range video comprehension, LVBench organizes evaluation around six core capabilities: entity recognition, event understanding, key information retrieval, temporal grounding, reasoning, and summarization. 
An additional strength of LVBench is its high-quality annotation protocol, where questions are manually designed to require visual evidence and are accompanied by temporal clues indicating the minimal relevant video segment. 
These properties make LVBench a challenging and representative benchmark for assessing multimodal models under extended temporal contexts.

Table~\ref{tab:lvbench} presents results on LVBench~\cite{wang2025lvbench}.
\method{} achieves an overall score of 52.9, outperforming all agent-based systems
that use open-source models, including MemVid (44.4), VCA (41.3), VideoTree (28.8),
and VideoAgent (29.3), and remaining competitive with the leading open-source MLLM
AdaReTaKe-72B (53.3) despite operating through a bounded agentic pipeline rather than
dense end-to-end inference.
VideoLucy (58.8) employs DeepSeek-R1~\cite{guo2025deepseek} as its agent base model
(denoted $\dagger$), yet \method{} narrows the gap to 5.9 points using only open-source
components throughout, demonstrating that our caption-once, frames-on-demand design
remains competitive even against pipelines backed by frontier closed-source reasoning models.
Among the six evaluation dimensions, \method{} achieves its strongest result on
Key Information Retrieval (63.9), where the dual-track narrative index provides
precisely timestamped textual anchors that facilitate accurate localization of
query-relevant content.

Notably, \method{} uses identical hyperparameters across all evaluated benchmarks
without any dataset-specific tuning, indicating that the caption-once dual-track
indexing and the visual-need routing strategy generalize robustly across
substantially different video durations and evaluation protocols,
from the short-to-long splits of Video-MME to the hour-long narratives of
LVBench and InfiniBench.

\begin{table*}[t]
\centering
\caption{Performance comparison on LVBench~\cite{wang2025lvbench} across six evaluation dimensions: 
\textit{Entity Recognition} (ER), \textit{Event Understanding} (EU), \textit{Key Information Retrieval} (KIR), \textit{Temporal Grounding} (TG), \textit{Reasoning} (Rea), and \textit{Summarization} (Sum). 
\textit{Overall} denotes the average score over all six dimensions. 
$\dagger$ denotes methods that employ a closed-source LLM as the agent base model.}
\setlength{\tabcolsep}{10pt} 
\resizebox{1\linewidth}{!}{%
\begin{tabular}{lccccccc}
\hline
Method & ER    & EU    & KIR   & TG    & Rea   & Sum   & Overall \bigstrut\\
\hline
\hline
\textit{Leading Open-source MLLMs} &       &       &       &       &       &       &  \bigstrut[t]\\
TimeMarker-8B~\cite{chen2024timemarker} & 42.8  & 39.1  & 34.9  & 38.7  & 38.2  & 48.8  & 41.3 \\
VideoLLaMA3-7B~\cite{zhang2025videollama} & 45.8  & 42.4  & 47.8  & 35.9  & 45.8  & 36.2  & 45.3 \\
InternVL2.5-78B~\cite{chen2024internvl25} & 43.8  & 42.0  & 42.1  & 36.8  & 51.0  & 37.9  & 43.6 \\
Qwen2-VL-72B~\cite{wang2024qwen2} & 38.0  & 41.1  & 38.3  & 41.4  & 46.5  & 46.6  & 41.3 \\
ReTake-7B~\cite{wang2024retake} & 49.8  & 46.2  & 52.9  & 45.0  & 45.8  & 27.6  & 47.8 \\
VideoChat-Flash-7B~\cite{li2024videochatflash} & 51.1  & 46.0  & 49.0  & 38.9  & 48.5  & 34.5  & 48.2 \\
AdaReTaKe-72B~\cite{wang2025adaretake} & 53.0  & 50.7  & 62.2  & 45.5  & 54.7  & 37.9  & 53.3 \\
InternVL3-8B~\cite{zhu2025internvl3} & 47.7  & 43.0  & 42.6  & 42.3  & 46.8  & 25.9  & 44.5 \bigstrut[b]\\
\hline
\textit{Agent-based Systems} &       &       &       &       &       &       &  \bigstrut[t]\\
VideoAgent~\cite{wang2024videoagent} & 28.0  & 30.3  & 28.0  & 29.3  & 28.0  & 36.4  & 29.3 \\
VideoTree~\cite{wang2025videotree} & 30.3  & 25.1  & 26.5  & 27.7  & 31.9  & 25.5  & 28.8 \\
MemVid~\cite{yuan2025memory} & 53.4  & 40.6  & 46.3  & 34.9  & 43.2  & 28.1  & 44.4 \\
VCA~\cite{yang2025vca} & 43.7  & 40.7  & 37.8  & 38.0  & 46.2  & 27.3  & 41.3 \\
\textcolor[rgb]{ .678,  .678,  .678}{VideoLucy~\cite{zuo2025videolucy}$^\dagger$} & \textcolor[rgb]{ .678,  .678,  .678}{54.3} & \textcolor[rgb]{ .678,  .678,  .678}{59.8} & \textcolor[rgb]{ .678,  .678,  .678}{75.6} & \textcolor[rgb]{ .678,  .678,  .678}{51.7} & \textcolor[rgb]{ .678,  .678,  .678}{55.9} & \textcolor[rgb]{ .678,  .678,  .678}{49.1} & \textcolor[rgb]{ .678,  .678,  .678}{58.8} \bigstrut[b]\\
\hline
\method{} (Ours) & 53.9  & 50.2  & 63.9  & 48.2  & 48.3  & 37.9  & 52.9 \bigstrut\\
\hline
\end{tabular}%

}
\label{tab:lvbench}
\end{table*}

\section{Further Ablation Study}
\label{sec:supp_ablation}

\noindent\textbf{Cloud reasoning model size.}
Table~\ref{tab:cloud_size} ablates the size of the cloud-side MLLM responsible for
the Answering Agent, Localization Agent, and Visual-Need Router.
Performance scales consistently with model size across all duration splits,
from 49.7 overall at 2B to 67.5 at 32B, confirming that reasoning quality and
routing accuracy are meaningful bottlenecks in the pipeline.
The gap between 2B and 8B (9.2 points overall) is substantially larger than that
between 8B and 32B (8.6 points), suggesting diminishing returns at the upper end.
In contrast to the captioner size ablation (Table~6 in the main paper), where scaling to 32B slightly degraded
long video performance due to caption verbosity accumulation, the cloud reasoning
model benefits monotonically from scale, as larger models provide more reliable
event localization and more accurate routing decisions without inflating the
narrative context.
We further observe that the 2B model frequently terminates at Step~1 by returning
a confident answer directly from the event-level story skeleton, bypassing the
localization and visual verification stages entirely.
This suggests that smaller models lack the calibration to assess their own uncertainty
reliably, and that effective agentic workflows with iterative backtracking currently
require sufficiently large reasoning models to engage the full pipeline.

\begin{table}[t]
\centering
\caption{Video-MME accuracy with different cloud reasoning model sizes.}
\resizebox{\linewidth}{!}{%
\begin{tabular}{lccccc}
\hline
Reasoning Model & Size  & short & medium & long  & overall \bigstrut\\
\hline
\hline
\multirow{3}[2]{*}{Qwen3-VL} & 2B    & 57.3  & 47.7  & 44.1  & 49.7 \bigstrut[t]\\
      & 8B    & 61.6  & 61.6  & 53.6  & 58.9 \\
      & 32B   & 72.2  & 66.6  & 63.6  & 67.5 \bigstrut[b]\\
\hline
\end{tabular}%

}
\label{tab:cloud_size}
\end{table}

\noindent\textbf{Generalization across model families.}
To assess the generalizability of \method{} beyond a single model family, we evaluate
two additional backbone configurations: InternVL3.5~\cite{wang2025internvl35}, a widely adopted open-source MLLM
family, and Qwen3.5~\cite{qwen2025qwen35}, a recently released series with stronger reasoning capabilities.
Table~\ref{tab:more_mllm} shows that \method{} with Qwen3-VL (8B captioner, 32B reasoner)
achieves 67.5 overall at 5.8 frames per question, while the InternVL3.5 configuration
(8B + 38B) reaches 64.3 at 7.6 frames, demonstrating that the pipeline transfers across
model families without architectural changes.
Replacing both the captioner and reasoner with Qwen3.5 (9B + 27B) further improves
overall performance to \textbf{69.4} with only \textbf{2.5 frames} per question,
outperforming all agent-based baselines and substantially closing the gap with
Qwen3-VL-32B (75.9) at 768 frames.
The lower frame count indicates that the Qwen3.5 backbone configuration triggers visual verification less frequently. Because both the captioner and reasoner are changed simultaneously, we do not attribute this reduction solely to routing quality; it may also reflect stronger textual memories, earlier confidence-based exits, or different localization behavior. Overall, these results demonstrate that CFD transfers across model families without architectural changes.
Notably, Qwen3.5 is a natively multimodal model that processes interleaved text and
visual inputs within a unified architecture, confirming that supplying a small number
of on-demand frames alongside rich textual memories at reasoning time is both
compatible with and well-suited to the emerging paradigm of native multimodal reasoning.

\begin{table*}[t]
\centering
\caption{Performance comparison on Video-MME~\cite{fu2025videomme} across multiple
backbone configurations.
$\dagger$~denotes reproduced results under strictly identical setups using the authors'
code.
\textit{\#Captions} denotes the average temporal span covered by each caption.
\textit{Re-caption?} indicates whether a method revisits video frames at query time
to (re)caption or refine its textual memory.
\textit{\#Frames} counts only raw frames sent to the online answering model.}
\resizebox{\linewidth}{!}{%
\begin{tabular}{lccccccc}
\hline
\multirow{2}[4]{*}{\textbf{Method}} & \multirow{2}[4]{*}{\textbf{\#Captions}} & \multirow{2}[4]{*}{\textbf{Re-caption?}} & \multirow{2}[4]{*}{\textbf{\#Frames}} & \multicolumn{4}{c}{\textbf{Video-MME}} \bigstrut\\
\cline{5-8}      &       &       &       & \textbf{short} & \textbf{medium} & \textbf{long} & \textbf{overall} \bigstrut\\
\hline
\hline
\textit{Leading Open-source MLLMs} &       &       &       &       &       &       &  \bigstrut[t]\\
VideoChat2-7B~\cite{li2024mvbench} & -     & -     & 16    & 48.3  & 37.0  & 33.2  & 39.5 \\
LongVA-7B~\cite{zhang2024longva} & -     & -     & 128   & 61.1  & 50.4  & 46.2  & 52.6 \\
Kangaroo-7B~\cite{liu2024kangaroo} & -     & -     & 64    & 66.1  & 55.3  & 46.6  & 56.0 \\
Video-CCAM-14B~\cite{fei2024video} & -     & -     & 96    & 62.2  & 50.6  & 46.7  & 53.2 \\
VideoXL-7B~\cite{shu2025videoxl} & -     & -     & 128   & 64.0  & 53.2  & 49.2  & 55.5 \\
Dispider-7B~\cite{qian2025dispider} & -     & -     & 1 fps & -     & -     & -     & 57.2 \\
VideoChat-Online-4B~\cite{huang2025online} & -     & -     & 2 fps & -     & -     & 47.1  & 54.4 \\
TimeChat-Online-7B~\cite{yao2025timechatonline} & -     & -     & 1 fps & -     & -     & 48.4  & 62.4 \\
Qwen3-VL-32B~\cite{bai2025qwen3vl} &       &       & 768   & 82.4  & 76.3  & 69.0  & 75.9 \bigstrut[b]\\
\hline
\textit{Agent-based Systems} &       &       &       &       &       &       &  \bigstrut[t]\\
VideoAgent~\cite{fan2024videoagent} & 2s / caption & \cmark & -     & -     & -     & 46.4  & - \\
VideoTree~\cite{wang2025videotree} & 8s / caption & \cmark & -     & 67.8  & 59.9  & 54.2  & 60.6 \\
DrVideo~\cite{ma2025drvideo} & 5s / caption & \cmark & -     & -     & -     & 51.7  & - \\
MemVid~\cite{yuan2025memory} & -     & \xmark & 1 fps & 73.9  & 63.1  & 55.0  & 64.0 \\
VideoLucy~\cite{zuo2025videolucy}$^\dagger$ & < 30s / caption & \cmark & -     & 73.2  & 64.7  & 56.2  & 64.7 \bigstrut[b]\\
\hline
\method{} (Qwen3-VL-8B + Qwen3-VL-32B) & 233.1s / caption & \xmark & 5.8   & 72.2  & 66.6  & 63.6  & 67.5 \bigstrut[t]\\
\method{} (InternVL3.5-8B + InternVL3.5-38B) & 233.1s / caption & \xmark & 7.6   & 69.0  & 63.1  & 60.7  & 64.3 \\
\method{} (Qwen3.5-9B + Qwen3.5-27B) & 233.1s / caption & \xmark & 2.5   & 73.7  & 70.4  & 64.1  & 69.4 \bigstrut[b]\\
\hline
\end{tabular}%

}
\label{tab:more_mllm}
\end{table*}

\begin{table}[t]
\centering
\caption{Online LLM serving time on Video-MME.}
\resizebox{\linewidth}{!}{%
\begin{tabular}{lc}
\hline
Policy & Online LLM serving time per query \bigstrut\\
\hline
\hline
No frames & 59.6 s \bigstrut[t]\\
Always frames & 63.3 s \\
Visual-Need Router & 68.0 s \bigstrut[b]\\
\hline
\end{tabular}%

}
\label{tab:latency}
\end{table}

\noindent\textbf{Reasoning efficiency.}
Table~\ref{tab:latency} reports online LLM serving time for no-frame,
always-frame, and routed inference. It is worth
noting that the Router adds an LLM call, so frame savings do not imply lower wall-clock latency in
every configuration.

\begin{figure*}[t]
\centering
\includegraphics[width=1\linewidth]{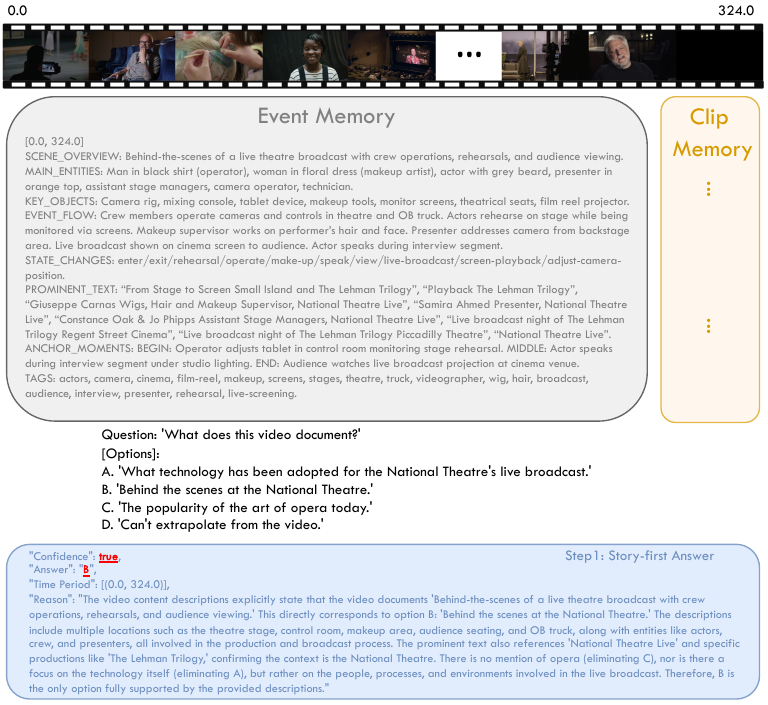}
\caption{A qualitative visualization of the direct-answer mode of \protect\method{} on a Video-MME example. Since the coarse event memory already provides sufficient global semantic evidence, the model answers correctly at the initial story-first stage without temporal backtracking or visual evidence retrieval.}
\label{fig:Visualization2} 
\end{figure*}

\noindent\textbf{Qualitative visualizations.}
Figure~\ref{fig:Visualization2} presents the complementary \emph{direct-answer} mode of \method{}, where the question can already be resolved from coarse event memory without any further backtracking. In this example, the retrieved event memory explicitly summarizes the video as a behind-the-scenes record of a live theatre broadcast, covering crew operations, rehearsals, makeup, control-room monitoring, and audience viewing, while the prominent text repeatedly anchors the scene to \textit{National Theatre Live}. As a result, the model answers correctly at the initial \textit{story-first} stage with high confidence, selecting option \textbf{B} directly from the global semantic storyline. Unlike the first case, this question does not depend on a fine-grained perceptual attribute or a narrowly localized visual cue; instead, it is fully supported by high-level narrative evidence already available in the event memory. This example therefore demonstrates the efficiency-oriented side of \method{}: when the memory abstraction is already sufficiently informative, the system stops early and avoids unnecessary event localization, routing, and visual evidence retrieval.

\begin{figure*}[t]
\centering
\includegraphics[width=1\linewidth]{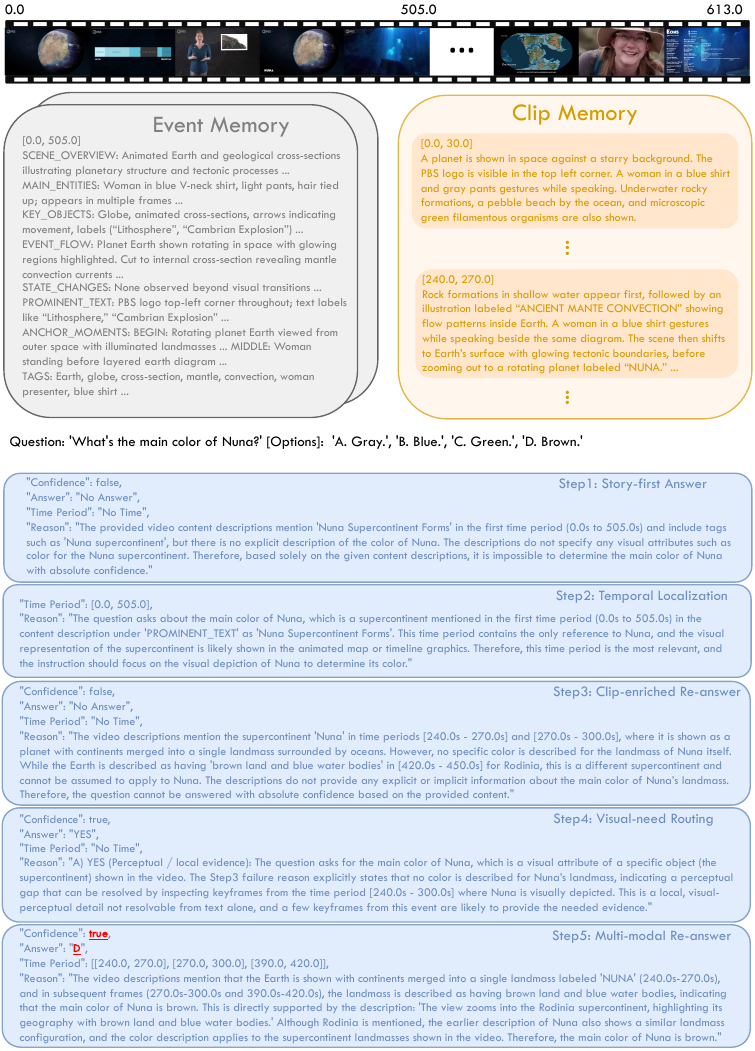}
\caption{A qualitative visualization of the inference trajectory of \protect\method{} on a Video-MME example. The model first reasons over coarse event memory, then progressively narrows the relevant temporal span with clip memory, and invokes sparse visual backtracking only when textual evidence remains insufficient.}
\label{fig:Visualization1} 
\end{figure*}

Figure~\ref{fig:Visualization1} visualizes the full inference trajectory of \method{} on a representative \textit{attribute perception} example, where the question asks for the main color of \textit{Nuna}. The case clearly illustrates the core design of our framework: \emph{story first, evidence on demand}. Starting from coarse \textit{Event Memory}, the model first attempts a story-level answer but correctly abstains, since the retrieved memory only indicates that ``Nuna Supercontinent Forms'' appears in the relevant event, while no explicit color cue is provided. It then performs temporal localization and selects the large event spanning [0.0, 505.0], after which a clip-enriched re-answer further narrows the relevant evidence to the local spans around [240,300] seconds, where \textit{NUNA} explicitly appears, yet still remains uncertain because the textual memory does not fully specify the target attribute. At this point, the Visual-Need Router correctly identifies the question as a \emph{local perceptual query}, rather than a long-horizon temporal reasoning problem, and triggers sparse evidence extraction only when textual memory is insufficient. The final multimodal re-answer then resolves the ambiguity by incorporating a small set of visual keyframes, leading to the correct prediction.

Together, the two examples highlight two important properties of \method{}: first, it avoids unnecessary visual access when high-level memory is already sufficient; second, when the missing information is fine-grained and perceptual, it can selectively backtrack to visual evidence instead of reprocessing the entire video.
Overall, the visualization demonstrates how our method progressively refines its evidence source from coarse narrative memory, to localized clip memory, and finally to sparse visual grounding, which is precisely the behavior needed for efficient long-video question answering.

\noindent\textbf{Practical motivation and cost accounting.}
A representative application is interactive wearable lifelogging. Commercial products such as Looki L1 capture daily experiences hands-free and transform them into lifelogs, stories, videos, and searchable memories. In our \emph{single-device, single-run hands-on test}, approximately 30 minutes of footage required around two hours to transfer from the device to the phone and then to the cloud, followed by more than five hours of preprocessing before AI interaction became available. We report this only as an illustrative observation under our specific device and network conditions, rather than as a general benchmark of the product. These observations motivate a clear separation between one-time indexing and interactive QA: CFD constructs the Event and Clip Memories once during offline indexing and reuses them across questions, so the cloud operates on the compact cached index by default and transfers only localized sparse keyframes when textual evidence is insufficient.

Table~\ref{tab:cost_accounting} makes this cost structure explicit, reporting offline captioning time, online serving time, cached-memory size, and GPU energy under the fixed eight-worker setup, using one H200 for Qwen3-VL-8B captioning and one H200 for Qwen3-VL-32B reasoning.

\begin{table*}[t]
\centering
\scriptsize
\caption{Offline/online cost accounting on Video-MME. Offline captioning is
one-time per video; online serving is per query. Energy integrates both H200
GPUs over one offline indexing pass and three Video-MME questions per video.}
\resizebox{\linewidth}{!}{%
\begin{tabular}{lcccc}
\hline
Metric & Short & Medium & Long & Average \\
\hline
\hline
Offline captioning (s/video) & 34.7 & 129.1 & 502.4 & 222.1 \\
Online LLM serving (s/query) & 39.3 & 66.7 & 97.9 & 68.0 \\
Cached caption memory (MB/video) & 0.0042 & 0.0191 & 0.0826 & 0.0353 \\
GPU energy (Wh/video + 3 queries) & 2.91 & 7.24 & 20.53 & 10.23 \\
Amortized time, $Q{=}1/3/5/10$ (s/query) &
74.0/50.9/46.2/42.8 &
195.8/109.7/92.5/79.6 &
600.3/265.3/198.3/148.1 &
290.0/142.0/112.4/90.2 \\
\hline
\end{tabular}

}
\label{tab:cost_accounting}
\end{table*}

The table separates the one-time offline cost from the per-query online cost. The amortized serving time is $T_{\mathrm{amortized}}(Q)=T_{\mathrm{offline}}/Q+T_{\mathrm{online}}$, where $Q$ is the number of questions sharing the same cached caption memories, $T_{\mathrm{offline}}$ is the one-time offline captioning time per video, and $T_{\mathrm{online}}$ is the online LLM serving time per query. Including preprocessing, cache writing, frame extraction, and orchestration, the corresponding end-to-end averages are \textbf{229.3 s/video offline} and \textbf{68.2 s/query online}; these end-to-end measurements are distinct from model-serving time. These measurements include server idle and serving overhead and should therefore be interpreted as conservative server-level measurements, not as measurements on a smartphone or wearable device.

\noindent\textbf{Visual-Need Router reliability.}
Since Video-MME has no routing annotations, we construct an \textbf{outcome oracle} from a paired diagnostic rerun over all 2,700 questions. The oracle uses frames only when the always-frame policy corrects an error made by the no-frame policy; it is an empirical upper bound rather than semantic ground-truth annotation. Table~\ref{tab:router_trigger} summarizes the Router's trigger rates and its coverage of frame-beneficial cases.

\begin{table*}[t]
\centering
\scriptsize
\caption{Router trigger statistics from the paired diagnostic rerun.}
\resizebox{0.8\linewidth}{!}{%
\begin{tabular}{lcccc}
\hline
Metric & Short & Medium & Long & Average \\
\hline
\hline
Queries retrieving frames & 34.1\% & 46.1\% & 58.3\% & 46.2\% \\
Frame-beneficial cases captured & 42/46 (91.3\%) & 46/47 (97.9\%) & 39/40 (97.5\%) & 127/133 (95.5\%) \\
\hline
\end{tabular}

}
\label{tab:router_trigger}
\end{table*}

\begin{table}[t]
\centering
\caption{Router decisions against the outcome oracle on the 1,876 decidable
questions in the paired diagnostic rerun.}
\resizebox{0.75\linewidth}{!}{%
\begin{tabular}{lcc}
\hline
 & Router YES & Router NO \\
\hline
\hline
Oracle YES & 127 & 6 \\
Oracle NO & 585 & 1,158 \\
\hline
\end{tabular}

}
\label{tab:router_confusion}
\end{table}

Oracle YES means that always-frame is correct while no-frame is wrong; Oracle NO means that no-frame is correct; cases where neither policy is correct are unresolved. Among 1,876 decidable questions, the confusion matrix gives \textbf{68.5\% routing accuracy} and \textbf{95.5\% recall} for frame-beneficial questions (Table~\ref{tab:router_confusion}). Table~\ref{tab:oracle_diagnostic} compares the outcome oracle with the no-frame, always-frame, and Router policies across duration splits.

\begin{table}[t]
\centering
\caption{Outcome-oracle analysis on Video-MME. The oracle is an empirical
upper bound, not a deployable policy. Each cell is (accuracy
 / online frames per query).}
\setlength{\tabcolsep}{3.5pt}
\resizebox{\linewidth}{!}{%
\begin{tabular}{lcccc}
\hline
Policy & Short   & Medium   & Long   & Average   \\
\hline
\hline
Outcome oracle & 74.0 / 0.33 & 69.9 / 0.73 & 64.6 / 0.87 & 69.5 / 0.64 \\
Visual-Need Router & 72.2 / 2.38 & 66.6 / 5.81 & 63.6 / 9.09 & 67.5 / 5.76 \\
\hline
\end{tabular}

}
\label{tab:oracle_diagnostic}
\end{table}

The comparison highlights the remaining failure modes: mixed temporal-perceptual questions, unnecessary routing caused by low textual confidence, and missed evidence after correct routing due to localization or sparse sampling. Confidence calibration, lightweight supervision, and learned routing policies are promising future directions.

\noindent\textbf{Evidence-depth allocation and perceptual diagnostics.}
The component and budget studies in the main paper establish the overall accuracy--cost trade-off; here we make explicit how the system allocates evidence at different depths. The caption-activation comparison in Table~\ref{tab:caption_reduction} shows that the all-clip baseline supplies all uniformly sampled 30-second Clip Memory captions to every query, whereas CFD activates only captions overlapping localized events. Text-only backtracking cannot recover evidence absent from the offline index, so this selective activation provides local detail without exposing the full clip memory. Event Memory uses TransNetV2 boundaries, while Clip Memory is generated uniformly and independently of event segmentation. The story-first loop then adapts reasoning depth to evidence sufficiency: simple global questions can terminate after coarse Event Memory, whereas harder questions progressively invoke temporal localization, localized Clip Memory, and, only when necessary, sparse visual evidence. Figures~\ref{fig:Visualization2} and \ref{fig:Visualization1} illustrate these two evidence-depth regimes.

The complete per-category Router statistics are reported in Table~\ref{tab:router_category}; routing is especially frequent for Attribute Perception and OCR, where pixel-level evidence is most likely to be decisive.

\begin{table}[t]
\centering
\caption{Category-level Router statistics from the paired diagnostic rerun.}
\resizebox{0.9\linewidth}{!}{%
\begin{tabular}{lccc}
\hline
Category & Router YES rate & Frames/query & Accuracy \\
\hline
\hline
Attribute Perception & 83.57\% & 4.02 & 77.93 \\
OCR Problems & 86.96\% & 3.37 & 82.73 \\
Temporal Reasoning & 69.76\% & 6.45 & 63.28 \\
Action Reasoning & 74.89\% & 9.51 & 58.25 \\
\hline
\end{tabular}

}
\label{tab:router_category}
\end{table}

\begin{table}[t]
\centering
\caption{Matched diagnostic comparison of text-only evidence and on-demand
pixels.}
\resizebox{0.8\linewidth}{!}{%
\begin{tabular}{lcc}
\hline
Category & Text-only & On-demand pixels \\
\hline
\hline
OCR & 77.0 & 82.7 \\
Spatial Perception & 70.4 & 75.9 \\
\hline
\end{tabular}

}
\label{tab:matched_perception}
\end{table}

Table~\ref{tab:matched_perception} reports the additional matched diagnostic comparison, in which on-demand pixels improve \textbf{76.98$\rightarrow$82.73} on OCR and \textbf{70.37$\rightarrow$75.93} on Spatial Perception.

\section{More Implementation Details}
\label{sec:impl}

\noindent\textbf{Captioning Agent.}
The offline Captioning Agent runs on the edge device and is invoked once per video
in a question-agnostic indexing pass.
It operates under two distinct prompt regimes depending on the memory tier being constructed.
For \emph{Event Memory}, the agent is prompted to produce a structured event profile
covering scene overview, entity listing, chronological event flow, state changes,
prominent text, and three retrieval anchor moments.
For \emph{Clip Memory}, the agent is prompted to produce plain-sentence micro-action
descriptions at high temporal granularity, one observable action per sentence,
with explicit instructions on object attribute inclusion, spatial
relations, and verbatim transcription of visible text.
The full prompts are provided in Tables~\ref{tab:prompt_event} and~\ref{tab:prompt_clip}.

\begin{table*}[ht]
\centering
\caption{Event Memory captioning prompt (Captioning Agent, offline).}
\label{tab:prompt_event}
\begin{tabularx}{\linewidth}{X}
\hline
\textbf{Goal} \\
You are an event-level video indexer.
The input is ONE event clip cropped from a longer video.
Your output will be used for: (1) retrieval / localization, (2) building a storyline skeleton,
(3) deciding what to inspect in finer memory. \\
\hline
\textbf{Rules} \\
- Describe ONLY what is directly observable in this event clip.
Do NOT guess names, roles (\eg, protagonist/antagonist), intent, emotions, relationships, or causes. \\
- Avoid subjective tone words (\eg, tense, dramatic, emotional, suspicious, implied). \\
- If something is uncertain, state it as ``unclear'' rather than guessing. \\
- Keep it compact: prefer short phrases over long sentences. \\
- If the event contains multiple distinct scenes/locations, explicitly represent the scene breaks and the chronological order. \\
\hline
\textbf{Output Format} (strict key-value; no extra text) \\
\texttt{SCENE\_OVERVIEW:} $\langle$1 sentence: dominant setting + time/lighting if visible$\rangle$ \\
\texttt{LOCATION\_SEQUENCE:} \\
\quad - $\langle$L1: place/area descriptor$\rangle$ \\
\quad - $\langle$L2: place/area descriptor$\rangle$ \ldots \\
\texttt{MAIN\_ENTITIES:} $\langle$comma-separated; include visible attributes: clothing color/type, notable features$\rangle$ \\
\texttt{KEY\_OBJECTS:} $\langle$comma-separated; only salient objects that appear or are interacted with$\rangle$ \\
\texttt{EVENT\_FLOW} (chronological phases; 4--8 bullets, each is ONE major beat): \\
\quad - $\langle$Phase 1: who + where + main visible action/change$\rangle$ \\
\quad - $\langle$Phase 2: \ldots$\rangle$ \\
\texttt{STATE\_CHANGES:} $\langle$opened/closed/on/off/enter/exit/sit/stand/hand-off etc.\ $\mid$ none$\rangle$ \\
\texttt{PROMINENT\_TEXT:} $\langle$verbatim visible overlay/sign/subtitle $\mid$ none $\mid$ unreadable$\rangle$ \\
\texttt{ANCHOR\_MOMENTS} (for retrieval; exactly 3): \\
\quad - BEGIN: $\langle$one short factual snapshot$\rangle$ \\
\quad - MIDDLE: $\langle$one short factual snapshot$\rangle$ \\
\quad - END: $\langle$one short factual snapshot$\rangle$ \\
\texttt{TAGS} (max 60; comma-separated): $\langle$only concrete nouns/short phrases: people attributes, objects, locations, visible texts, core actions; no full sentences$\rangle$ \\
\hline
\textbf{User turn:} Write the event profile now. \\
\hline
\end{tabularx}
\end{table*}

\begin{table*}[ht]
\centering
\caption{Clip Memory captioning prompt (Captioning Agent, offline).}
\label{tab:prompt_clip}
\begin{tabularx}{\linewidth}{X}
\hline
\textbf{1) Task Description} \\
You are a precise visual captioner for micro-detail video logging.
The input is one short clip cropped from a longer video.
Your output will be used for retrieval, evidence alignment, and answering questions later. \\
\hline
\textbf{2) Instructions} \\
- Output ONLY plain sentences. No headers, no bullet points, no structured keys. \\
- Describe ONLY what is directly observable. Do not guess intent, emotions, names, or unseen causes. \\
- POV rule: if the video is clearly first-person POV, use first-person (``I\ldots''); otherwise use third-person (``A person\ldots'', ``The man\ldots'', ``The camera\ldots''). \\
- Granularity rule: exactly ONE micro-action per sentence. If nothing moves, write one sentence describing the stable state (who/what/where). \\
- Information density: include object attributes (color/type/shape), spatial relations (left/right/on/in/near/behind), and state changes (opened/closed/on/off) whenever visible.
If any text or symbol appears, transcribe it verbatim. If text is present but unreadable, write ``text unreadable''. \\
- Keep sentences short and factual. Prefer concrete nouns over pronouns when possible. \\
\hline
\textbf{3) Examples} \\
\textit{I am in a kitchen. I pick up a red mug from the table.} \\
\textit{A man stands by a white refrigerator. He opens the refrigerator door.} \\
\hline
\textbf{User turn:} Caption this clip: \\
\hline
\end{tabularx}
\end{table*}

\begin{table*}[h]
\centering
\caption{Prompt for the Answering Agent. Given the current narrative memory and
optional visual evidence, the agent determines whether sufficient information
exists to answer confidently.}
\label{tab:prompt_answer}
\begin{tabularx}{\linewidth}{X}
\hline
The following provides a rough description of what's shown in the video during
different time periods: \textit{\{Current Memory List (Time Period + Content Description)\}} \\[4pt]
Note that since these descriptions are not very complete and detailed, some key
information in the video segments of each time period may not all appear in these
content descriptions. \\[4pt]
Now, a question has been raised regarding the content descriptions of this video.
\textit{\{Question and Options\}} \\[4pt]
Please read the given video content descriptions and the question in depth, and
determine whether you can accurately answer the given question solely based on
the currently provided descriptions.
If you can answer it with absolute confidence, please answer this question and
provide the time periods you are referring to.
The answer you provide must have completely and absolutely objective support in the
video descriptions.
Do not make inferences arbitrarily.
Please note that there is only one option that can answer this question.
If you think the current content descriptions of the video are still insufficient
to accurately answer the question, please do not answer it and give me your reason. \\[4pt]
You must note that if an ordinal number appears in the provided question,
in the vast majority of cases, you should not simply assume that this ordinal number
represents the ordinal of the provided time period.
You need to focus on understanding the specific meaning of this ordinal number within
the question based on all the content descriptions. \\[4pt]
Output in a strictly standardized dictionary with four key-value pairs: \\
\texttt{"Confidence"}: boolean. \texttt{True} if certain; \texttt{False} if not. \\
\texttt{"Answer"}: the answer option letter when confident; \texttt{"No Answer"} otherwise. \\
\texttt{"Time Period"}: list of \texttt{(start, end)} tuples when confident; \texttt{"No Time"} otherwise. \\
\texttt{"Reason"}: reasoning that absolutely supports the answer.
No additional comments should be added within the dictionary. \\
\hline
\end{tabularx}
\end{table*}

\begin{table*}[h]
\centering
\caption{Prompt for the Localization Agent. Given the current narrative memory and
the set of already-explored events, the agent identifies the single most
question-relevant unexplored time period.}
\label{tab:prompt_loc}
\begin{tabularx}{\linewidth}{X}
\hline
There is currently a video with a total duration of \textit{\{video length\}} seconds.
The following gives a general description of what is shown in the video during certain
time periods: \textit{\{Current Memory List (Time Period + Content Description)\}} \\[4pt]
Now, a question has been raised regarding this entire video.
\textit{\{Question and Options\}} \\[4pt]
Please read the given video content descriptions and the question in depth.
You do not need to answer this question. \\[4pt]
Your task is to identify, based on the video content in each time period,
the single time period that is most relevant to the question and that you think
requires further elaboration of its video content details to make the answer
to this question more explicit.
Notably, you only need to select the most relevant one from the time periods other
than the following already-explored periods:
\textit{\{Already Searched Time Periods\}} \\[4pt]
Output in a strictly standardized dictionary with two key-value pairs: \\
\texttt{"Time Period"}: list containing the single most relevant \texttt{(start, end)}. \\
\texttt{"Reason"}: justification for the selected time period.
No additional comments should be added within the dictionary. \\
\hline
\end{tabularx}
\end{table*}

\begin{table*}[h]
\centering
\caption{Prompt for the Visual-Need Router. Given the current selected event,
the question, and the Step~3 failure reason, the router decides whether sparse
keyframes from the selected event will improve correctness.}
\label{tab:prompt_router}
\begin{tabularx}{\linewidth}{X}
\hline
You are a routing agent for an agentic multimodal long video QA system.
We already attempted answering using text memories (EventMemory + injected ClipMemory
captions), but confidence was low.
Keyframes are sparse local evidence; captions are better for temporal abstraction
(scene transitions, long-horizon unfolding).
Now decide whether to extract a few visual keyframes from the \textbf{current selected event}.
Important: you are not deciding whether video is useful in general; you are deciding
whether keyframes from \emph{this} event will likely improve correctness for
\emph{this} question. \\[4pt]
\textbf{Current Event:} from \textit{\{start\}}s to \textit{\{end\}}s.
Event Text Memory: \textit{\{Event Memory\}} \\
\textbf{Step~3 Failure Reason:} \textit{\{Reason\}} \\
\textbf{Question:} \textit{\{Question and Options\}} \\[4pt]
\textbf{A) Return YES}, the missing evidence is visual-perceptual and likely
visible in a few keyframes from this event: appearance (clothing, colors, patterns);
object attributes (type/shape/color); visible text, signs, overlays, logos;
spatial/state cues (left/right, holding, on/in/near, open/closed, on/off);
micro-action disambiguation within this event (which hand/tool, exact contact,
small gesture).
Step~3 reason indicates ambiguity that a look can resolve (cannot distinguish options
based on text). \\[4pt]
\textbf{B) Return NO}, the question requires temporal abstraction where
captions are more suitable: temporal context focus (how things evolve over time,
what happens first/then/later, timeline/unfolding, duration-based reasoning);
scene transitions (when/where the scene changes, cut/transition/montage, switching
locations/activities across a long segment); multi-stage processes within a long span
where captions already summarize phases better than a few frames; global narrative
linking multiple events or summarization-level questions.
Step~3 reason indicates need for broader temporal coverage (earlier/later parts,
multiple moments) rather than local visual detail. \\[4pt]
\textbf{C) Keyword heuristics} (weak signals):
lean YES: color, wearing, looks like, logo, label, text says/reads, number on,
left/right, holding, where is the object.
Lean NO: scene change/transition/cut, timeline, unfold, sequence, earlier/later,
before/after (across moments), how long, over time, montage. \\[4pt]
\textbf{D) Default policy} (type-driven):
if the question is primarily perceptual-local $\rightarrow$ YES;
if primarily temporal-structural / scene-transition $\rightarrow$ NO;
if mixed, follow Step~3 failure reason: perceptual uncertainty $\rightarrow$ YES,
temporal coverage uncertainty $\rightarrow$ NO. \\[4pt]
Output a valid JSON object with exactly four keys: \\
\texttt{"Confidence"}: boolean. \\
\texttt{"Answer"}: \texttt{"YES"} or \texttt{"NO"}. \\
\texttt{"Time Period"}: \texttt{"No Time"}. \\
\texttt{"Reason"}: short justification referencing the rubric (A/B/D). \\
\hline
\end{tabularx}
\end{table*}

\noindent\textbf{Answering Agent.}
Following prior agent-based video QA systems~\cite{zuo2025videolucy}, the Answering
Agent receives the current narrative memory assembled from Event Memory with
selectively injected Clip Memory entries, and optionally the visual evidence
in the FIFO working memory $\mathcal{W}$, assembled into a single multimodal prompt.
Given this context and the question, the agent determines whether the available
evidence is sufficient to answer with confidence.
It outputs a structured dictionary with four fields: \texttt{Confidence} (boolean),
\texttt{Answer} (the selected option letter when confident, \texttt{"No Answer"}
otherwise), \texttt{Time Period} (supporting time intervals when confident,
\texttt{"No Time"} otherwise), and \texttt{Reason} (a chain-of-thought justification
that must provide absolute evidential support for the answer).
The full prompt is provided in Table~\ref{tab:prompt_answer}.

\noindent\textbf{Localization Agent.}
Following~\cite{zuo2025videolucy}, the Localization Agent identifies the single
most question-relevant unexplored event from the current narrative memory,
excluding all time periods already visited in the current query session.
It outputs the selected time period with reasons.
The full prompt is provided in Table~\ref{tab:prompt_loc}.

\noindent\textbf{Visual-Need Router.}
After a text-only answer attempt fails to reach confidence, the router receives
the selected event, the question, and the Step~3 failure reason, and decides
whether sparse keyframes from the current event are likely to improve correctness.
The routing decision follows a perceptual-versus-temporal rubric: it returns
\texttt{YES} when the missing evidence is visual-perceptual in nature, such as
object appearance, clothing color, visible text, spatial layout, or fine-grained
state cues that cannot be reliably inferred from text descriptions alone;
it returns \texttt{NO} when the question primarily concerns temporal structure,
scene transitions, or long-horizon narrative unfolding, where the textual memory
already provides a more suitable abstraction than sparse frames.
The full prompt is provided in Table~\ref{tab:prompt_router}.

\newpage

\end{document}